\documentclass{article}

\PassOptionsToPackage{numbers,square,sort&compress}{natbib}
\usepackage[main,final]{neurips_2026}

\usepackage[utf8]{inputenc}
\usepackage[T1]{fontenc}
\usepackage[hypertexnames=false,hidelinks]{hyperref}
\usepackage{url}
\usepackage{graphicx}
\usepackage{booktabs}
\usepackage{amsmath,amssymb,amsfonts}
\usepackage{nicefrac}
\usepackage{microtype}
\usepackage{xcolor}
\usepackage{colortbl}
\usepackage{enumitem}
\usepackage{multirow}
\usepackage{pifont}
\usepackage{float}
\usepackage{adjustbox}
\usepackage[capitalize]{cleveref}
\usepackage[normalem]{ulem}
\usepackage{tabularx}
\usepackage{soul}
\usepackage{bbm}
\usepackage{wrapfig}
\usepackage{makecell}
\usepackage{capt-of}
\usepackage{algorithm}
\usepackage{algorithmic}

\usepackage[accsupp]{axessibility}

\usepackage{xspace}
\makeatletter
\DeclareRobustCommand\onedot{\futurelet\@let@token\@onedot}
\def\@onedot{\ifx\@let@token.\else.\null\fi\xspace}

\makeatother

\definecolor{policyblue}{RGB}{230,242,255}

\definecolor{P1red}{RGB}{220,38,38}        
\definecolor{P2orange}{RGB}{234,138,0}     
\definecolor{P3blue}{RGB}{37,99,235}       
\definecolor{guidegray}{RGB}{100,100,100}  
\newcommand{\Pcritical}[1]{\textcolor{P1red}{\textbf{[P1-CRITICAL: #1]}}}
\newcommand{\Pmajor}[1]{\textcolor{P2orange}{\textbf{[P2-MAJOR: #1]}}}

\newcommand{\guide}[1]{\par\noindent\textcolor{guidegray}{\small\textit{Guidance: #1}}\par}

\newcommand{\method}{2CAP}

\newcommand{\cmark}{\ding{51}}
\newcommand{\xmark}{\ding{55}}
\newcommand{\Ecam}{\mathcal{E}_{\mathrm{cam}}}
\newcommand{\Esem}{\mathcal{E}_{\mathrm{sem}}}

\newcommand{\xq}{\mathbf{x}_q}
\newcommand{\xref}{\mathbf{x}_{\mathrm{ref}}}

\newcommand{\scam}{s_{\mathrm{cam}}}
\newcommand{\ssem}{s_{\mathrm{sem}}}

\newcommand{\TBD}[1]{{\color{red}\textbf{#1}}}

\crefname{section}{Sec.}{Secs.}
\Crefname{section}{Section}{Sections}
\crefname{table}{Tab.}{Tabs.}
\Crefname{table}{Table}{Tables}

\newcommand{\StageAPairs}{80{,}465}
\newcommand{\StageADresden}{17{,}242}
\newcommand{\StageASIDCamera}{15{,}000}
\newcommand{\StageAUnpairedCamera}{32{,}242}
\newcommand{\StageAUnpairedEdited}{15{,}000}
\newcommand{\StageAEntries}{208{,}172}

\begin{document}

\title{%
  \smash{\raisebox{-0.3em}{\includegraphics[height=1.6em,trim=40 40 40 50,clip]{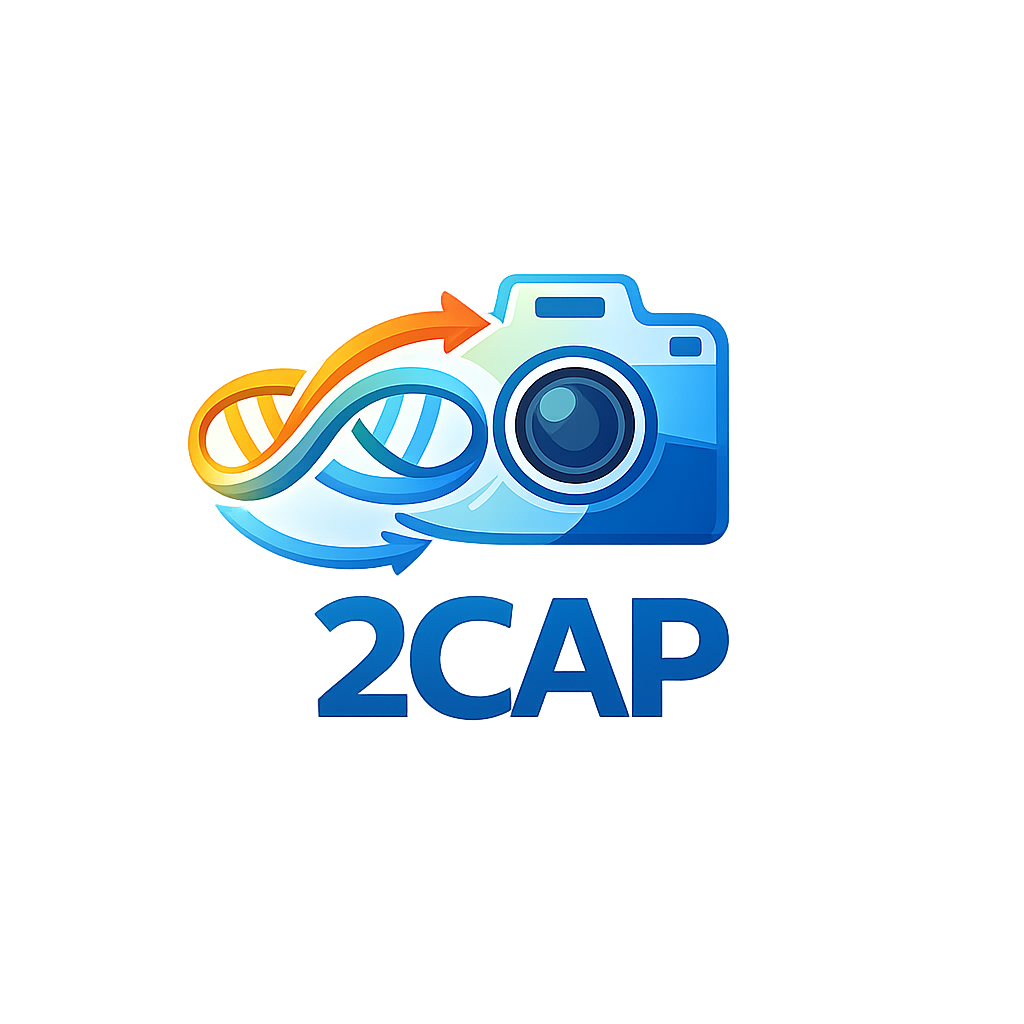}}}\hspace{-0.2em}%
  Beyond Real or Fake:\\
  A Dual-Channel Authenticity and Reasoning Protocol for Photographic Assessment%
}

\author{%
  Xiaoxiao Li\textsuperscript{1,3,4} \qquad Ruinan Jin\textsuperscript{1,3,4} \qquad Lili Meng\textsuperscript{1} \\
  \textbf{Miaosen Wang\textsuperscript{2} \qquad Athula Balachandran\textsuperscript{1} \qquad Pei Cao\textsuperscript{1}} \\
  \normalfont\textsuperscript{1}Google \qquad \textsuperscript{2}Google DeepMind \qquad \textsuperscript{3}University of British Columbia \qquad \textsuperscript{4}Vector Institute
}
\maketitle

\begin{abstract}
Forensic verification has converged on a binary ``real vs.\ fake'' label that conflates fully synthetic, tampered, and AI-retouched images despite their very different consequences. We argue that GenAI manipulations decompose along two complementary channels: a \textit{camera channel} of sensor fingerprints (destroyed by neural processing), and a \textit{semantic channel} of scene-level coherence (disrupted by content edits). Each manipulation type leaves a distinctive two-channel signature. We instantiate this taxonomy in \textbf{\method} (\textbf{2}-\textbf{C}hannel \textbf{A}uthenticity \textbf{P}rotocol), pairing a contrastively-trained camera encoder with a frozen semantic encoder to serve two applications on a shared backbone: \textbf{(i)} four-class authenticity classification via cross-attention fusion with learnable reliability weights; and \textbf{(ii)} evidence generation, where an authentic image serves as a reference for a query and the two encoders produce per-channel alignment scores together with a patch-level saliency map. These signals are supplied as privileged context to a frozen Vision-Language Model (VLM) through an Observe-Generate-Refine loop, yielding explanations with localization without forensic fine-tuning. On a multi-source benchmark, \method{} attains the best classification performance, including strong retouching detection where existing methods fail, and improves explanation quality.
\end{abstract}

\section{Introduction}
\label{sec:intro}
\begin{figure}[t]
    \centering
    \includegraphics[width=0.99\textwidth]{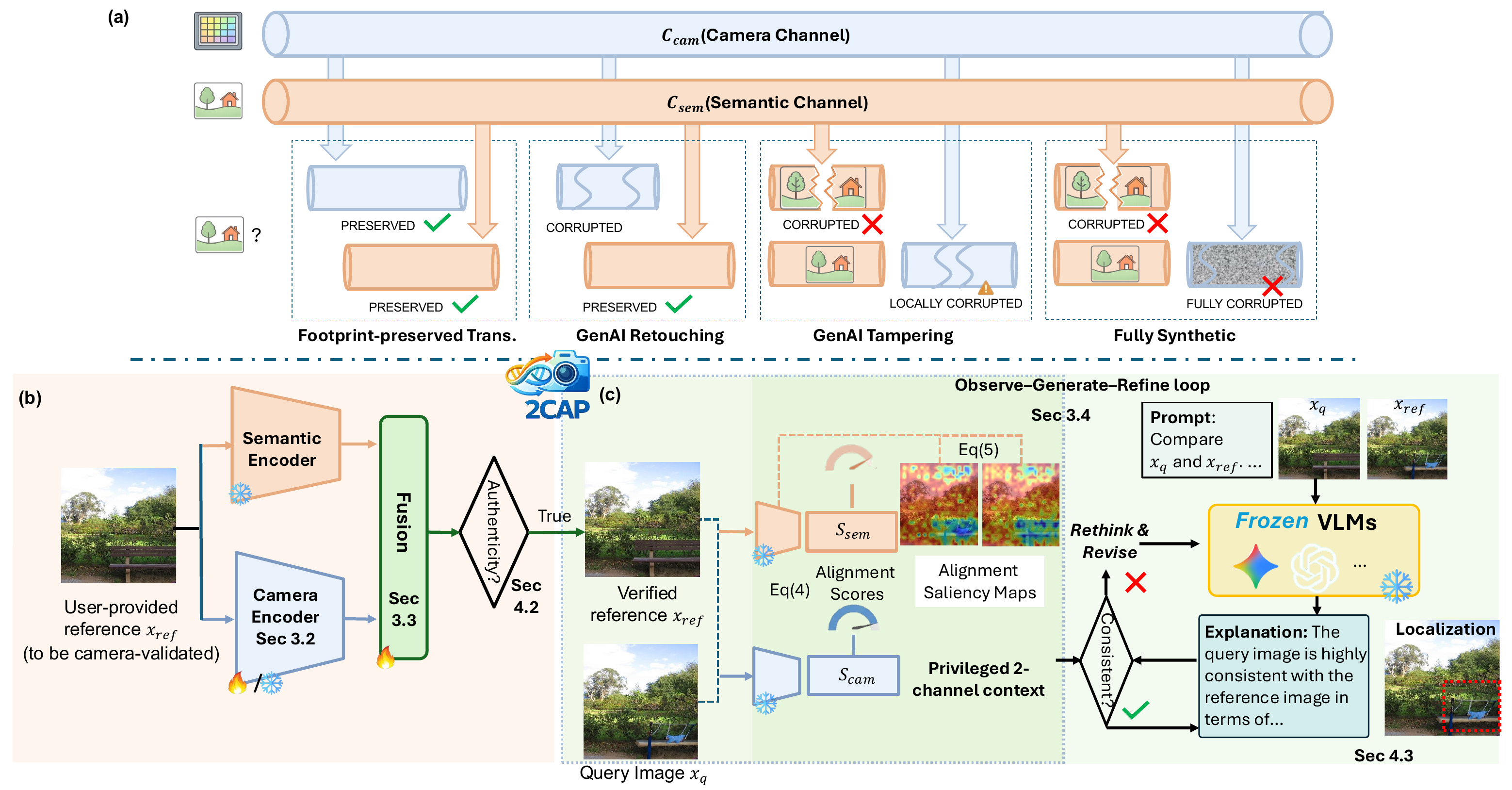}
    \caption{\textbf{\method{} overview.} (a)~\textit{Two-channel taxonomy.} Authenticity cues are decoupled into a camera channel $C_{\mathrm{cam}}$ (sensor traces) and a semantic channel $C_{\mathrm{sem}}$ (scene coherence); each manipulation type leaves a distinct signature: footprint-preserved transformations preserve both channels, GenAI retouching corrupts only $C_{\mathrm{cam}}$, GenAI tampering corrupts $C_{\mathrm{cam}}$ globally and $C_{\mathrm{sem}}$ locally, and fully synthetic images corrupt both. (b)~\textit{Authenticity classification} (Q1). A camera encoder (Sec.~\ref{sec:camera_encoder}) and a frozen semantic encoder feed a cross-attention fusion module with learnable reliability weights (Sec.~\ref{sec:fusion}) that produces a four-class verdict (Sec.~\ref{sec:four_class}). (c)~\textit{Reference-based evidence generation} (Q2). Given a query $\xq$ and a camera-verified reference $\xref$, the two encoders yield channel alignment scores $\scam, \ssem$ (Eq.~\ref{eq:scores}) and a patch-level saliency map (Eq.~\ref{eq:saliency}); these serve as privileged two-channel context to a frozen VLM in an Observe-Generate-Refine loop (Sec.~\ref{sec:vlm}) that returns an explanation and localization, revising whenever the VLM's initial assessment is inconsistent with the channel evidence (Sec.~\ref{sec:vlm_eval}).}
    \label{fig:architecture}
    \vspace{-4mm}
\end{figure}

When generative AI (GenAI) tools modify a photorealistic image, they do not all break it the same way. A portrait restyled by an enhancement model~\cite{rombach2022high,podell2023sdxl} may still depict the same person while altering the original capture traces; a news photograph with an inserted object~\citep{brooks2023instructpix2pix} may retain camera-related cues in unedited regions while changing local semantic content; a fully synthesized scene~\citep{flux2024} has no camera capture of the depicted scene, yet may remain semantically coherent. These scenarios can demand different responses, which a binary authenticity label alone does not express.

Cryptographic provenance (C2PA~\citep{c2pa2024}) addresses authenticity at capture time but cannot reach legacy or stripped content. Computational forensics fills this gap, but the field has organized around binary classification, across both forgery localization~\citep{wu2019mantra,dong2022mvss,guillaro2023trufor} and AI-generated image detection~\citep{ojha2023towards,wang2020cnn,yan2024sanity,liu2024forgery,zhang2025towards}, grouping AI-generated, edited, and retouched images under a single ``fake'' label regardless of how or why they were modified. This loses the structured evidence that journalism, legal proceedings, identity verification, and content moderation actually need: \textit{what was changed, and does it matter under a given policy?} For example, camera-parameter retouching may restyle photographic appearance while preserving the people, objects, and scene relationships relevant to a factual claim. Such an image is modified, but the modification alone does not establish that its depicted information is false; a blanket ``fake'' label therefore conflates processing history with factual fidelity. Whether the edit is acceptable remains dependent on the claim, context, and applicable policy. We argue that useful forensic systems must not only detect modifications, but characterize GenAI modification type by distinguishing (i)~\textit{retouching} (camera-parameter restyling intended to preserve scene content), (ii)~\textit{tampering} (localized semantic edits), and (iii)~\textit{full synthesis} (no camera capture).

We organize photographic assessment around two complementary evidence channels: a \textbf{camera channel} sensitive to capture and processing cues, and a \textbf{semantic channel} encoding scene content and relationships. We adopt camera-captured, GenAI-retouched, locally tampered, and fully synthetic images as four prototypical image-formation and editing regimes (Fig.~\ref{fig:architecture}(a)). They distinguish questions that binary real/fake detection conflates: whether an image originates from camera capture, whether its photographic appearance has been restyled, and whether its scene content has been modified or generated. These distinctions define an operational benchmark for evaluating the information contributed by the two channels across forms of image creation and editing.

Practical workflows also compose operations: local content editing followed by denoising and recompression can affect both semantic correspondence and camera-related traces. We therefore use the four regimes as reference cases for studying complementary evidence, rather than as an exhaustive partition of processing histories. Channel responses can overlap: benign processing can weaken capture cues, and synthesis can remain semantically coherent. Continuous query-reference alignment measurements describe such responses relative to a corresponding reference, while four-class prediction remains a dataset-defined task on joint image features.

We propose \textbf{\method} (\textbf{2}-\textbf{C}hannel \textbf{A}uthenticity \textbf{P}rotocol), which implements this evidence-based formulation with two encoders feeding two complementary applications. The shared backbone consists of a contrastively-trained camera encoder that maps images into a learned space of camera-related processing cues, paired with a frozen SigLIP2~\citep{tschannen2025siglip2} semantic encoder. \textbf{(1)~Reference-free authenticity classification} (Fig.~\ref{fig:architecture}(b)): a cross-attention fusion module with learnable reliability weights adaptively combines the two channels per single image to produce a four-class verdict over camera, retouched, tampered, and fully synthetic. \textbf{(2)~Reference-based evidence generation} (Fig.~\ref{fig:architecture}(c)): given a query image and a trusted corresponding reference, we treat per-channel alignment scores and a patch-level saliency map as \textit{privileged context} that a frozen VLM cannot compute on its own, and feed them through a structured Observe-Generate-Refine (OGR) loop based on the iterative self-refinement paradigm of \citet{madaan2023selfrefine} to obtain an explanation and a localized bounding box. 

\noindent\textbf{Contributions.}
\textbf{(1)~A new problem formulation.} We move image forensics beyond binary real/fake classification by organizing assessment around camera-related and semantic evidence across four prototypical regimes. We evaluate four-class discrimination using the joint features and inspect pairwise channel scores for camera, retouched, and tampered images.
\textbf{(2)~A new detection capability.} We instantiate this formulation in \method{}, a cross-attention fusion framework with learnable reliability weights that achieves the highest overall AUC and F1 among the compared methods on our benchmark. 
\textbf{(3)~Forensic explanation without VLM fine-tuning.} We show that \textit{frozen} VLMs can produce structured forensic evidence without any forensic SFT data: channel similarity scores and saliency maps, provided as privileged context through an OGR loop, improve change-type accuracy over the single-pass baseline across four evaluated VLMs, with model- and metric-dependent effects in the component ablations.

\section{Related Work}
\label{sec:related}

\noindent\textbf{Image forgery detection and localization.}
Classical forensics exploit sensor fingerprints~\citep{lukas2006digital,chen2008determining} and compression traces~\citep{bianchi2012image,amerini2014splicing}. Learning-based methods fuse noise and RGB cues for end-to-end detection and localization~\citep{wu2019mantra,dong2022mvss,guillaro2023trufor,cozzolino2020noiseprint}. In parallel, AI-generated image detection improves cross-generator generalization via pretrained representations~\citep{wang2020cnn,ojha2023towards,tan2024frequency,liu2024forgery,zhang2025towards,yan2024sanity,yang2025d}. These methods largely target binary detection or localization; our evaluation additionally distinguishes GenAI retouching as a separate category. Extended discussion is in Appendix~\ref{app:relatedwork}.

\noindent\textbf{Training data and representation controls.} B-Free~\citep{guillaro2025bias} reduces content bias through aligned real/synthetic training data, while DRCT~\citep{chen2024drct} uses diffusion-reconstructed hard examples with contrastive training. SSAFE~\citep{lee2026ssafe} studies detection with frozen multimodal encoders. These methods motivate separating the effects of training data, representation choice, and fusion architecture when interpreting our results.

\noindent\textbf{VLM-based forensic reasoning.}
Recent methods train VLMs for forensic detection, localization, and explanation, including FakeVLM~\citep{wen2025spot}, FakeShield~\citep{xu2024fakeshield}, ForgeryGPT~\citep{liu2024forgerygpt}, and SIDA~\citep{huang2025sida}. SIDA predicts tamper masks as well as explanations, while its 300K-image dataset should not be equated with 300K explanation-instruction pairs. More recent work explores expert-grounded reasoning through trained models, including REVEAL~\citep{cao2025reveal}. Our distinct evaluation concerns whether channel evidence improves a frozen VLM's reference-based assessment. We demonstrate this across four proprietary configurations across two model families (GPT-4o and Gemini~2.5), without forensic instruction tuning of the VLM; this does not imply that the encoders are training-free or that OGR outperforms fine-tuned forensic VLMs.

\section{Method}
\label{sec:method}

\subsection{Verification Protocol and Problem Formulation}
\label{sec:protocol}

We consider two practical forensic scenarios with complementary input requirements:
\begin{enumerate}[leftmargin=1.5em,itemsep=1pt,topsep=3pt]
  \item \textbf{Q1: Is a given image authentic?} Given a single image $\mathbf{x}$, predict its dataset-defined label: camera-captured, GenAI-retouched, locally tampered, or fully synthetic. This is a \textit{reference-free} (single-image) task.
  \item \textbf{Q2: What changed between two images?} Given a query $\xq$ and a reference $\xref$ (e.g., a verified camera image), identify the nature, type, and location of modifications. 
\end{enumerate}

The reference-free classifier predicts a benchmark label from the joint encoder features; it does not infer an exhaustive processing history. For reference-based assessment, the continuous camera and semantic alignment scores $(\scam,\ssem)$ provide a two-dimensional description of evidence relative to a corresponding reference. This separates benchmark classification from the interpretation of combined processing effects: overlapping scores need not imply identical histories, and a composite workflow need not correspond uniquely to one of the four reference regimes.

\noindent\textbf{Two applications on a shared dual-channel foundation.} Both questions are answered by combining two complementary encoders: a camera encoder $\Ecam$ (Sec.~\ref{sec:camera_encoder}) that learns camera-related processing cues, and a frozen semantic encoder $\Esem$ (SigLIP2~\citep{tschannen2025siglip2}) that captures scene-level coherence. On top of these shared dual-channel features we build two applications, evaluated independently in Sec.~\ref{sec:experiments}. \textbf{(1)~Authenticity classification} (Sec.~\ref{sec:fusion}; evaluated in Sec.~\ref{sec:four_class}): a cross-attention fusion module with learnable reliability weights adaptively combines the two channels per single input to produce a four-class verdict. \textbf{(2)~VLM evidence generation} (Sec.~\ref{sec:vlm}; evaluated in Sec.~\ref{sec:vlm_eval}): per-channel alignment scores between $\xq$ and $\xref$, together with a patch-level saliency map, are supplied as privileged context to a frozen VLM through an Observe-Generate-Refine loop, producing interpretable evidence with localization.


\subsection{Camera Fingerprint Encoder}
\label{sec:camera_encoder}

\noindent\textbf{Limitation of existing pretrained encoders.}
Pretrained semantic encoders provide useful forensic features~\citep{yang2025d,liu2024forgery}, but their training objective does not explicitly target capture and processing traces. We therefore complement frozen semantic features with a camera-related encoder trained on camera/edit examples, and evaluate their contributions separately.

\begin{figure}[t]
    \centering
    \includegraphics[width=0.99\textwidth]{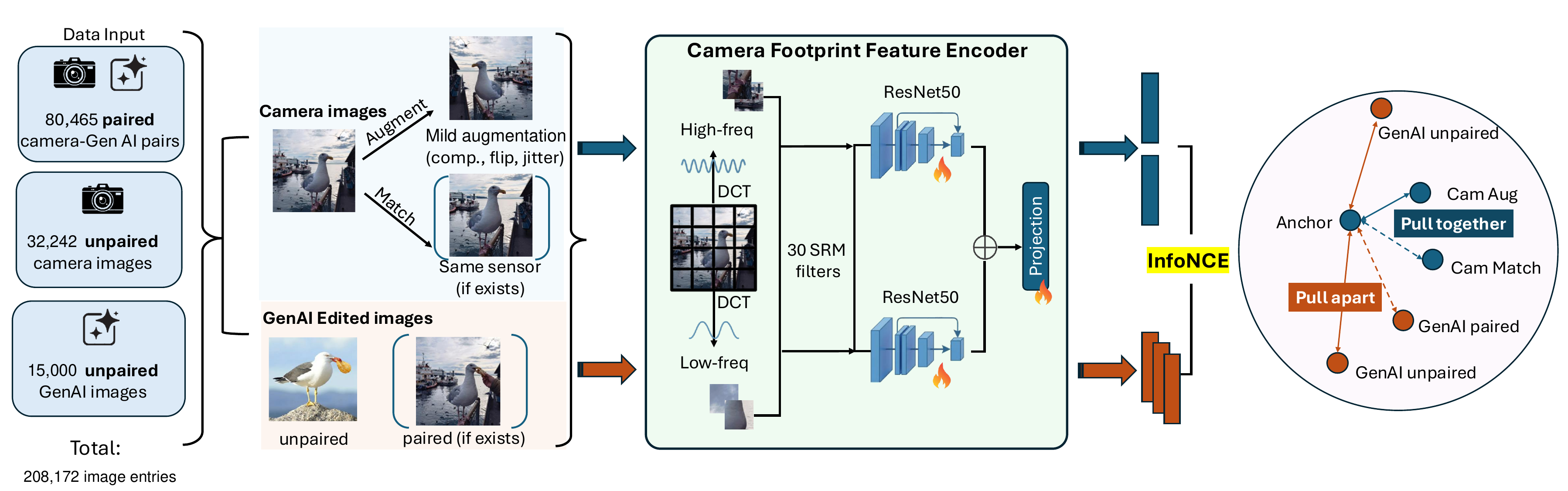}
    \caption{\textbf{Camera encoder pretraining.} \textit{Left:} Training data composition: \StageAPairs{} pairs, \StageAUnpairedCamera{} unpaired camera entries, and \StageAUnpairedEdited{} unpaired edited entries, totaling \StageAEntries{} image entries (Table~\ref{tab:stage1_data}); unique images have not been deduplicated. \textit{Center:} DCT-domain encoder with frequency-based patch selection and dual-branch feature extraction. \textit{Right:} InfoNCE training where camera image embeddings are pulled together while edited image embeddings are pushed apart.}
    \label{fig:stage_a}
    \vspace{-2mm}
\end{figure}

\noindent\textbf{Camera encoder architecture.}
We adopt AIDE's DCT+SRM dual-branch backbone~\citep{yan2024sanity} and reprogram its training objective from cross-entropy classification to InfoNCE contrastive metric learning (Figure~\ref{fig:stage_a}). The low-level feature extraction pipeline (DCT-based patch ranking and 30 SRM steganalysis filters) is inherited; our contribution is the training scheme, the cross-channel fusion, and the frozen-VLM evidence generation (a full architectural comparison is in Appendix~\ref{app:aide_diff}). Specifically, (i)~we train with InfoNCE contrastive learning over camera-camera positives and paired or unpaired camera-edit negatives, producing a 128-d unit-norm embedding space rather than class logits; (ii)~we use cosine similarity in this space as a continuous camera-channel score $\scam$ rather than a binary decision; and (iii)~we fuse camera features with features from a frozen semantic encoder (Sec.~\ref{sec:fusion}). 


\noindent\textbf{Contrastive pretraining for generalizability.}
A key design principle is that our camera encoder should learn \textit{camera-specific traces}, not semantic shortcuts that fail to generalize across generators or editing tools. We encourage this through contrastive learning (Figure~\ref{fig:stage_a}).

We construct two types of positives for each camera anchor: 1) \textit{Camera Aug}: A mild augmentation of the same image (horizontal flip, mild color jitter, and mild JPEG compression). 2) \textit{Camera Match}: Another image from the same camera sensor, when available. This encourages the encoder to learn sensor-consistent representations.
The negative pool consists of all edited images in the batch, including 1) \textit{GenAI paired}: The GenAI-edited version of the anchor's source image. This provides hard negatives that share scene content but lack authentic camera traces; and 2) \textit{GenAI unpaired}: Other edited images in the batch, providing diverse negative examples across different editing styles and generators. The total loss combines paired and unpaired streams:
\begin{equation}
  \mathcal{L}_{\mathrm{total}} = \mathcal{L}_{\mathrm{paired}} + \lambda\, \mathcal{L}_{\mathrm{unpaired}},
\label{eq:total_loss}
\end{equation}
where $\lambda = 0.5$. Each term is an InfoNCE loss with temperatures $\tau_p = 0.07$ (paired) and $\tau_u = 0.1$ (unpaired). 

\subsection{Cross-Channel Fusion for Authenticity Classification}
\label{sec:fusion}

The camera and semantic encoders provide features with different training objectives. We combine them using bidirectional cross-attention and input-dependent gating (Appendix~\ref{app:architecture}). The complementarity of the learned representations is evaluated through channel and fusion ablations, rather than assumed from a deterministic mapping between manipulation type and channel response.

\noindent\textbf{Cross-attention fusion.} We distinguish the token features used for classification from the global embeddings used for pairwise cosine scores. Let $Z_c\in\mathbb{R}^{B\times N_c\times d_c}$, for $c\in\{\mathrm{cam},\mathrm{sem}\}$, denote the channel patch-token sequences, where $N_{\mathrm{cam}}$ and $N_{\mathrm{sem}}$ are the numbers of camera and semantic patches (tensor shapes are summarized in Table~\ref{tab:cross-architecture}). Learned projections map them to $\widetilde Z_c\in\mathbb{R}^{B\times N_c\times256}$. For each attention head,
\begin{equation}
 \operatorname{Attn}(Q,K,V)=\operatorname{softmax}_{\mathrm{keys}}\!\left(QK^{\mathsf T}/\sqrt{d_h}\right)V,
\end{equation}
with camera-to-semantic attention weights of shape $B\times N_{\mathrm{cam}}\times N_{\mathrm{sem}}$ and the reverse shape in the other direction. Residual cross-attention yields $Z'_{\mathrm{cam}}$ and $Z'_{\mathrm{sem}}$; channel-wise pooling produces the 256-d vectors $\mathbf z'_c$ used below. Attention is normalized over the key patches for each query patch, producing a patch-to-patch interaction matrix rather than a single attention coefficient.

\noindent\textbf{Reliability-weighted combination.} Not all channels are equally informative for every image. A two-layer gating MLP takes the concatenated channel vectors and produces two logits,
\[
 \boldsymbol{\ell}=g_{\mathrm{gate}}([\mathbf z'_{\mathrm{cam}};\mathbf z'_{\mathrm{sem}}])\in\mathbb R^2,
 \qquad
 \alpha_c=\frac{\exp(\ell_c)}{\exp(\ell_{\mathrm{cam}})+\exp(\ell_{\mathrm{sem}})},
 \quad c\in\{\mathrm{cam},\mathrm{sem}\}.
\]
Here, softmax is applied over the two channels for each image, so $\alpha_{\mathrm{cam}}+\alpha_{\mathrm{sem}}=1$. This normalization is separate from the key-patch softmax inside cross-attention. We fuse the channel vectors using these input-dependent weights:
\begin{equation}
    \mathbf{z}_{\mathrm{fused}} = \alpha_{\mathrm{cam}} \cdot \mathbf{z}_{\mathrm{cam}}' + \alpha_{\mathrm{sem}} \cdot \mathbf{z}_{\mathrm{sem}}', \quad
    \mathbf p = \mathrm{softmax}(h_{\mathrm{cls}}(\mathbf{z}_{\mathrm{fused}}))
\end{equation}

The classifier $h_{\mathrm{cls}}$ has dimensions $256\rightarrow128\rightarrow4$ and is trained with four-class cross-entropy; its output categories are camera-captured, retouched, tampered, and fully synthetic.

\subsection{VLM Evidence Synthesis with Privileged-Context Feedback}
\label{sec:vlm}


The dual-channel classifier (Sections~\ref{sec:camera_encoder}-\ref{sec:fusion}) answers Question~1 (camera authenticity). To answer Question~2 (what changed) with interpretable evidence, we employ a structured Observe-Generate-Refine (OGR) loop with frozen VLMs, following the iterative self-refinement paradigm~\citep{madaan2023selfrefine} and taking the \textit{image pair} as input for direct comparison.

\noindent\textbf{Channel alignment scores and saliency.} We compute alignment scores via cosine similarity $\mathrm{sim}(\cdot,\cdot)$:
\begin{equation}
  \scam = \mathrm{sim}\!\left(\Ecam^{g}(\xq),\, \Ecam^{g}(\xref)\right), \quad
  \ssem = \mathrm{sim}\!\left(\Esem^{g}(\xq),\, \Esem^{g}(\xref)\right)
\label{eq:scores}
\end{equation}
where $\mathcal E_c^{g}$ denotes the global image embedding of channel $c$, distinct from its patch-token output $Z_c=\mathcal E_c^{\mathrm{tok}}(\mathbf x)$ used for cross-attention. The camera global embedding is 128-d and unit-norm. These are learned feature similarities, not probabilities of authenticity or direct measurements of retained sensor information. Low $\scam$ can reflect processing beyond GenAI, while high $\ssem$ indicates similar global content without excluding small local edits. For spatial localization, we compute a saliency map using patch-level features $\{p_i(x)\}$ from SigLIP2~\citep{tschannen2025siglip2}:
\begin{equation}
  \mathcal{M}_{\mathrm{sem},i} = 1 - \max_{j} \mathrm{sim}\!\left(p_i(\xq),\, p_j(\xref)\right),
\label{eq:saliency}
\end{equation}
where high $\mathcal{M}_{\mathrm{sem},i}$ indicates query patches lacking close matches in the reference. The resulting map $\mathcal{M}_{\mathrm{sem}} = \{\mathcal{M}_{\mathrm{sem},i}\}$ highlights query patches with weak reference matches. This directed map can miss removals or repeated-object changes when remaining query patches still find close matches.

\noindent\textbf{Motivation.}
A frozen VLM comparing two images tends to describe visually salient differences first (lighting, color) and may miss two important cases: (i)~retouching that changes camera traces without visible content differences, and (ii)~localized edits in low-saliency regions. Channel scores quantify evidence the VLM cannot directly compute (e.g., sensor-trace similarity), so providing them as feedback after an initial assessment lets the VLM correct both failure modes.

\noindent\textbf{OGR protocol.} Unlike prior methods that train VLMs for forensic tasks~\citep{wen2025spot,huang2025sida,liu2024forgerygpt}, we keep the VLM frozen and supply pre-computed signals ($\scam$, $\ssem$, $\mathcal{M}_{\mathrm{sem}}$) through a two-pass Observe-Generate-Refine loop (Algorithm~\ref{alg:ogr}). \textit{Pass~1 (Generate):} the VLM receives the image pair and outputs an initial assessment including change type (addition/removal/replacement), description, and bounding box. \textit{Pass~2 (Refine):} the VLM receives its own Pass-1 output together with the channel scores and saliency map, and revises if the evidence conflicts with its initial assessment. The VLM is not forced to agree with the scores; it performs independent reasoning and can confirm, refine, or override. Revisions can correct prior errors or introduce new ones; accuracy contrasts alone do not identify the two transition counts. We use a single prompt template across the tested VLM configurations; Appendix~\ref{app:qual_extended} shows examples of the structured outputs. Evaluation across four frozen VLMs (Table~\ref{tab:ogr-main-results}) shows that the evidence contribution depends on both model and metric.

\begin{algorithm}[t]
\caption{OGR Loop: Observe-Generate-Refine}
\label{alg:ogr}
\begin{algorithmic}[1]
\REQUIRE Image pair $(\xq, \xref)$, scores $(\scam, \ssem)$, saliency $\mathcal{M}_{\mathrm{sem}}$
\STATE \textbf{Observe:} VLM receives $(\xq, \xref)$ and performs visual inventory
\STATE \textbf{Generate:} VLM outputs initial assessment: change type, description, bounding box
\STATE \textbf{Refine:} VLM receives Pass-1 output with $(\scam, \ssem, \mathcal{M}_{\mathrm{sem}})$; revises if evidence conflicts
\RETURN Structured evidence: $\{\hat{y}, \scam, \ssem, \mathcal{M}_{\mathrm{sem}}, \text{explanation}\}$
\end{algorithmic}
\end{algorithm}

\section{Experiments}
\label{sec:experiments}
Our experiments evaluate the two applications of \method{}'s dual-channel features, mirroring the questions posed in Sec.~\ref{sec:intro}. \textbf{(i)~Dual-channel classifier} (Sec.~\ref{sec:four_class}): a reference-free four-class classifier with cross-attention fusion and learnable reliability weights, addressing Q1 (\textit{is the image authentic?}). \textbf{(ii)~OGR feedback loop} (Sec.~\ref{sec:vlm_eval}): a privileged-context protocol that supplies similarity scores and  saliency map to frozen VLMs for evidence generation, addressing Q2 (\textit{what changed?}). 

\subsection{Dataset Construction and Implementation Details}
\label{sec:dataset}

We construct a multi-source benchmark for distinguishing \textit{camera-captured} images from \textit{edited} images across three modification categories (see Appendix~\ref{sec:appendix_dataset_details} for full details):

\begin{itemize}[leftmargin=1.5em,itemsep=1pt,topsep=2pt]
  \item \textbf{Fully synthetic}: Images generated entirely by GenAI from text prompts, lacking camera fingerprints, including  Defactify Image (MS~COCOAI)~\citep{roy2026comprehensive} (SD2.1~\citep{rombach2022high}, SDXL~\citep{podell2023sdxl}, SD3~\citep{esser2024scaling}, DALL-E~3~\citep{ramesh2021zero}, Midjourney~6~\footnote{\url{https://www.midjourney.com/home}}) and the Flickr8K-derived subset of RU-AI~\citep{huang2025ru} (SDv1.5~\citep{rombach2022high}, SDAbsoluteReality~\footnote{\url{https://huggingface.co/Lykon/AbsoluteReality}}, SDepiCRealism~\footnote{\url{https://huggingface.co/emilianJR/epiCRealism}}, SDXL~\citep{podell2023sdxl}, SDv6.0~\citep{ren2024hyper}).
  \item \textbf{Tampered}: Localized GenAI editing (object insertion/removal/inpainting) that corrupts semantic content in specific regions. Sources: SID-Set~\citep{huang2025sida}, SAGI-D~\citep{giakoumoglou2025sagi}, and MagicBrush~\citep{zhang2023magicbrush}.
  \item \textbf{GenAI-retouched}: Generative Photography~\citep{yuan2025generative} dataset with global camera-parameter restyling (bokeh, color temperature, focal length, motion blur) using AnimateDiff~\citep{guo2023animatediff} intended to preserve scene content.
\end{itemize}

\noindent Camera images are aggregated from Dresden~\citep{gloe2010dresden}, VISION~\citep{shullani2017vision}, Defactify Image~\citep{roy2026comprehensive}, and SID-Set~\citep{huang2025sida}; paired camera originals from SAGI-D and MagicBrush also contribute to contrastive pretraining (Table~\ref{tab:stage1_data}). This multi-source design broadens coverage, but unequal source and tool distributions across classes can introduce confounding. Shared sources across some classes do not eliminate that risk.

\noindent\textbf{Implementation Details.} Our architecture comprises a camera encoder $\Ecam$ (DCT-domain dual-branch ResNet-50~\citep{he2016deep,yan2024sanity} with SRM filters, outputting 128-d embeddings), a frozen semantic encoder $\Esem$ (SigLIP2~\citep{tschannen2025siglip2}, 768-d), and a cross-attention fusion module with learnable reliability weights. We pretrain $\Ecam$ on \StageAEntries{} image entries using InfoNCE contrastive learning (Table~\ref{tab:stage1_data}).

\subsection{Camera Authenticity: Classification on Camera vs. Fully Synthetic vs. Tampered vs. GenAI-retouched Images}
\label{sec:four_class}


This section evaluates the \textit{first} application: the dual-channel classifier with cross-attention fusion and learnable reliability weights (Sec.~\ref{sec:fusion}), trained to assign each image to one of four classes: \textit{camera-captured}, \textit{fully synthetic}, \textit{tampered}, or \textit{GenAI-retouched}, the categories corresponding to distinct forensic scenarios requiring different responses.

\subsubsection{Baselines and Setup}
We compare against ten methods: TruFor~\citep{guillaro2023trufor}, D3~\citep{yang2025d}, CNNSpot~\citep{wang2020cnn}, AIDE~\citep{yan2024sanity}, VIBAIGCD~\citep{zhang2025towards}, FreqNet~\citep{tan2024frequency}, FatFormer~\citep{liu2024forgery}, UnivFD~\citep{ojha2023towards}, SAFE~\citep{li2025improving}, and GAPL~\citep{qin2026scaling}.

\noindent\textbf{Training protocol.} The adapted baseline rows use the historical 12,800-image downstream training pool. For the controlled \method{} ablation, we exclude 468 training rows whose source-reference groups overlap the historical test split, deduplicate four repeated training paths, and then partition the remainder into 9,877 training and 2,451 validation images by source-reference group. All six \method{} variants use these same examples, frozen encoder embeddings, minibatch order, optimizer settings, and 20-epoch budget; validation macro-F1 selects each checkpoint before one evaluation on the common 3,200-image test split (800 images per class). The full-model row uses a separate rerun in which all weights shared with the no-gate control start identically and the gate starts at equal weights; the no-gate control is the previously completed run under the same seed and recipe. Thus the method and baseline rows share the test population but \emph{not} the exact downstream training population or adaptation budget. \method{} also uses separate task-specific contrastive pretraining, so Table~\ref{tab:four_class_results} should not be read as a fully matched baseline comparison. The historical test split was inspected during prior development and is not a pristine blind holdout.

\noindent\textbf{Evaluation protocol.} VISION and the Flickr8K-derived RU-AI subset are held out from both of our training stages. This tests dataset and camera-domain transfer. Dataset holdout does not itself establish generator-family holdout: for example, SDXL appears in both the listed training-source and held-out-source generator sets. We distinguish these settings from the stricter evaluation in which an entire generator or editing-tool family is excluded from all training and model selection.

\subsubsection{Results}
\noindent\textbf{Quantitative classification results.}  A key contribution of \method{} is distinguishing \textit{among} modification types. All evaluations in this section are \textit{reference-free} (single-image). 
Table~\ref{tab:four_class_results} shows that the paired-initialization \method{} run achieves overall AUC .982 and F1 .906 on the common four-class test split. These exceed the reported baseline values, subject to the downstream-training differences described above.
For retouching on GP, \method{} reaches OvR AUC .998 and F1 .964, compared with F1 .868 for D3, the strongest compared baseline. The semantic-only variant reaches retouching F1 .914; the full model is 5.0 points higher and exceeds semantic-only overall F1 by 12.5 points. These differences support combining both channels on this test population, but do not isolate a benefit from reliability gating. Generalization is setting-dependent: Flickr8K AUC is .859 for the full model, below FatFormer (.938) and GAPL (.996).


\noindent\textbf{Ablation.} Table~\ref{tab:four_class_results} (bottom) compares the six variants under the common training and selection protocol. \emph{Channels.} Semantic-only obtains AUC .938 / F1 .781 and camera-only .939 / .775, below the full model (.982 / .906). \emph{Fusion strategy.} Concat-MLP reaches .977 / .890 and addition-MLP .978 / .888. Cross-attention with fixed equal channel weights reaches .983 / .908, slightly \emph{above} the reliability-gated model (.982 / .906). The full and no-gate variants have matched initial shared weights; in this single-seed comparison, the learned gate has no demonstrated performance benefit. 


\begin{table}[t]
\centering
\caption{\textbf{Four-class authenticity classification.} Overall and per-class performance on the common 3,200-image test set, plus held-out datasets ($^*$). Per-class metrics report one-vs-rest AUC (OvR) and F1. Best in \textbf{bold}, second \ul{underlined} (ranking computed over baselines and \method{} only). The bottom block uses a common group-filtered training/validation protocol for all six variants and is excluded from the ranking; the full row uses paired, neutral gate initialization and baseline training populations differ.}
\label{tab:four_class_results}
\scriptsize
\setlength{\tabcolsep}{2.2pt}
\renewcommand{\arraystretch}{1.08}
\resizebox{\columnwidth}{!}{%
\begin{tabular}{@{}lcccccccccccccc@{}}
\toprule
\textbf{Method} &
\multicolumn{4}{c}{\textbf{Overall}} &
\multicolumn{2}{c}{\textbf{Camera}} &
\multicolumn{2}{c}{\textbf{Tampered}} &
\multicolumn{2}{c}{\textbf{Synthetic}} &
\multicolumn{2}{c}{\textbf{Retouched}} &
\textbf{Flickr8K$^*$} &
\textbf{VISION$^*$} \\
\cmidrule(lr){2-5}
\cmidrule(lr){6-7}
\cmidrule(lr){8-9}
\cmidrule(lr){10-11}
\cmidrule(lr){12-13}
\cmidrule(lr){14-14}
\cmidrule(lr){15-15}
& \textbf{Prec.} & \textbf{Rec.} & \textbf{F1} & \textbf{AUC}
& \textbf{OvR} & \textbf{F1}
& \textbf{OvR} & \textbf{F1}
& \textbf{OvR} & \textbf{F1}
& \textbf{OvR} & \textbf{F1}
& \textbf{AUC}
& \textbf{Brier} $\downarrow$ \\
\midrule
TruFor~\citep{guillaro2023trufor}
& .100 & .254 & .123 & .559 & .618 & .410 & .724 & .000 & .497 & .000 & .396 & .082 & .637 & \ul{.123} \\
D3~\citep{yang2025d}
& .755 & .754 & .750 & .928 & .864 & .591 & .899 & .680 & .973 & .863 & \ul{.975} & \ul{.868} & .671 & .179 \\
CNNSpot~\citep{wang2020cnn}
& .370 & .361 & .359 & .631 & .601 & .349 & .565 & .316 & .685 & .430 & .672 & .344 & .583 & .237 \\
AIDE~\citep{yan2024sanity}
& \ul{.767} & \ul{.770} & \ul{.767} & \ul{.934} & \ul{.872} & \ul{.638} & \ul{.946} & \ul{.777} & .963 & .835 & .952 & .816 & .870 & .228 \\
VIBAIGCD~\citep{zhang2025towards}
& .752 & .740 & .730 & .925 & .863 & .523 & .890 & .673 & \ul{.975} & \ul{.871} & .974 & .851 & .879 & .253 \\
FreqNet~\citep{tan2024frequency}
& .530 & .518 & .484 & .768 & .663 & .258 & .791 & .569 & .886 & .688 & .731 & .422 & .840 & .329 \\
FatFormer~\citep{liu2024forgery}
& .539 & .547 & .537 & .783 & .695 & .422 & .727 & .452 & .851 & .621 & .858 & .653 & \ul{.938} & .159 \\
UnivFD~\citep{ojha2023towards}
& .380 & .323 & .269 & .615 & .576 & .415 & .559 & .061 & .600 & .252 & .727 & .347 & .818 & .165 \\
SAFE~\citep{li2025improving}
& .591 & .594 & .583 & .838 & .776 & .464 & .891 & .669 & .888 & .675 & .797 & .523 & .937 & .136 \\
GAPL~\citep{qin2026scaling}
& .455 & .445 & .387 & .695 & .796 & .548 & .637 & .396 & .727 & .524 & .620 & .081 & \textbf{.996} & \textbf{.095} \\
\midrule
\textbf{\method{}}
& \textbf{.908} & \textbf{.907} & \textbf{.906} & \textbf{.982} & \textbf{.957} & \textbf{.836} & \textbf{.987} & \textbf{.899} & \textbf{.987} & \textbf{.926} & \textbf{.998} & \textbf{.964} & .859 & .394 \\
\midrule
\multicolumn{15}{@{}l}{\textit{\color{gray}Ablation (ours): channels}} \\
 \quad Semantic only ($\Esem$)
  & .779 & .784 & .781 & .938 & .875 & .628 & .915 & .707 & .973 & .874 & .990 & .914 &
   .850 & .634 \\
 \quad Camera only ($\Ecam$)
  & .782 & .776 & .775 & .939 & .905 & .706 & .960 & .811 & .960 & .815 & .933 & .767 &
   .944 & .270 \\
\multicolumn{15}{@{}l}{\textit{\color{gray}Ablation (ours): fusion strategy}} \\
 \quad Concat + MLP
  & .890 & .890 & .890 & .977 & .947 & .821 & .982 & .886 & .985 & .914 & .994 & .938 &
  .914 & .464 \\
  \quad Addition + MLP
  & .888 & .888 & .888 & .978 & .948 & .818 & .983 & .881 & .986 & .915 & .995 & .939 &
  .928 & .390 \\
  \quad Cross-Attn (no rel.\ wts)
  & .908 & .908 & .908 & .983 & .958 & .834 & .987 & .899 & .989 & .931 & .998 & .968 &
  .914 & .281 \\
\bottomrule
\end{tabular}%
}
\end{table}

\begin{figure}[t]
\centering
\begin{minipage}[c]{0.43\linewidth}
  \centering
  \includegraphics[width=\linewidth]{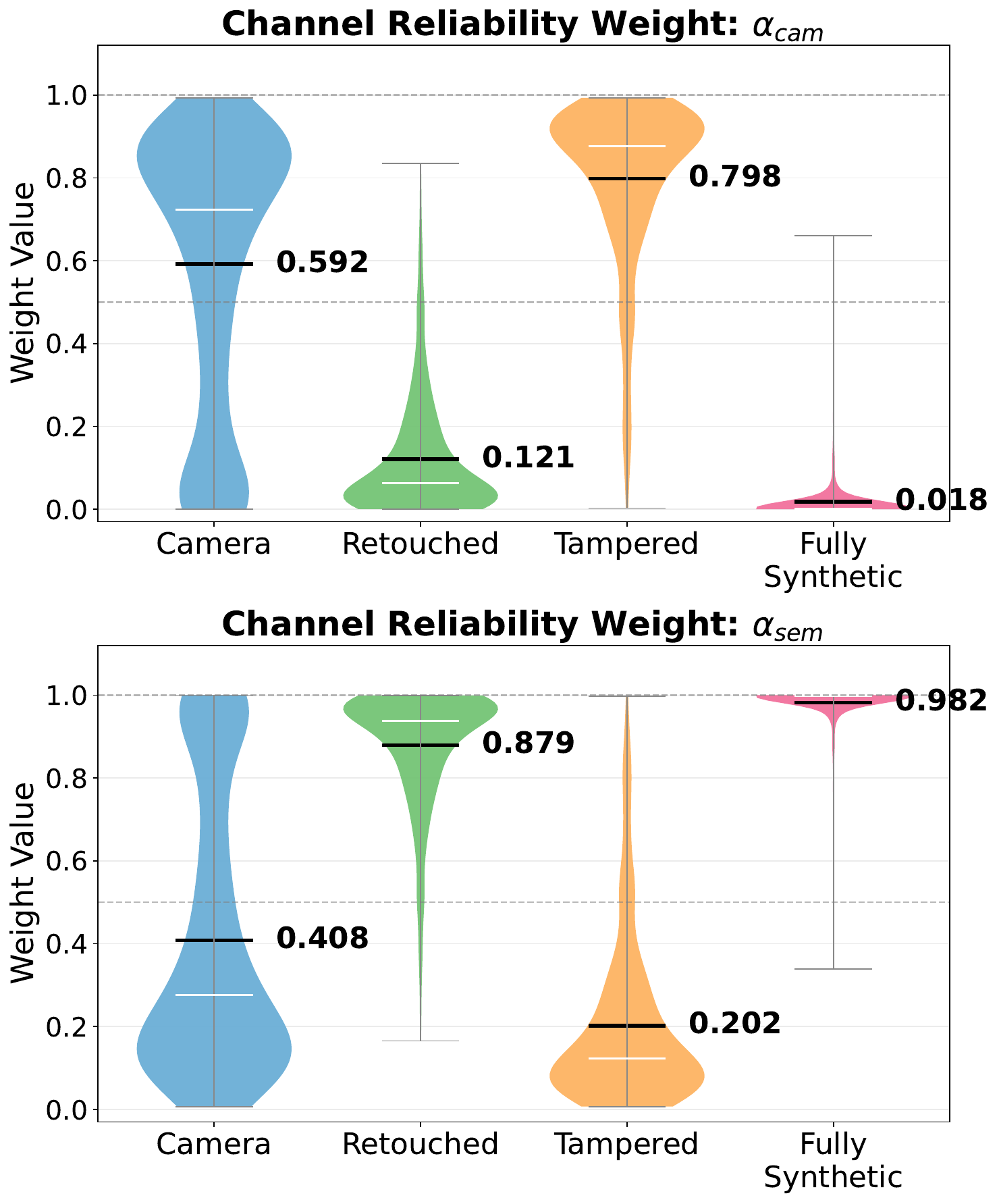}
\end{minipage}%
\hfill
\begin{minipage}[c]{0.57\linewidth}
  \centering
  \includegraphics[width=\linewidth]{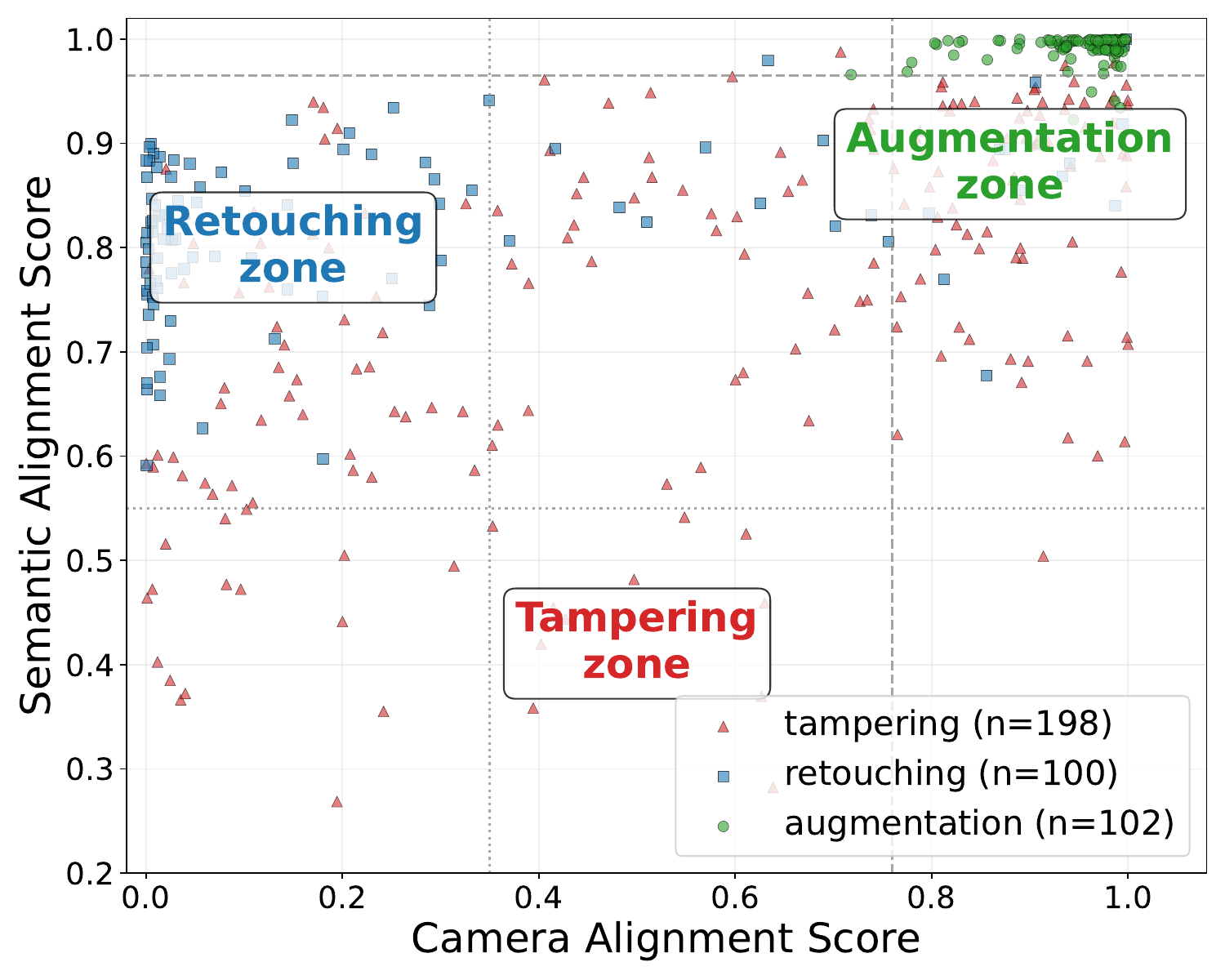}
\end{minipage}
\caption{\textbf{Left: learned reliability weights} $\alpha_{\mathrm{cam}}$ and $\alpha_{\mathrm{sem}}$ shown by category as descriptive post-attention diagnostics. \textbf{Right: score distribution.} Retouching: low~$\scam$, high~$\ssem$; camera: high/high; tampering: higher variance.}
\label{fig:score-scatter}
\label{fig:alpha-dist}
\vspace{-3mm}
\end{figure}

\noindent\textbf{Learned reliability weights.} Fig.~\ref{fig:alpha-dist} (left) shows category-dependent gate distributions: the camera-side representation has higher weight for tampered images, whereas the semantic-side representation is favored for retouching and synthesis. These are descriptive diagnostics, not calibrated uncertainty or causal feature attributions. Each representation has already received information from the other channel through cross-attention; gate magnitudes alone therefore do not establish which original forensic cue caused a prediction.

\subsection{Evaluation on VLM Evidence Generation}
\label{sec:vlm_eval}

\begin{figure}[t]
    \centering
\includegraphics[width=0.99\linewidth]{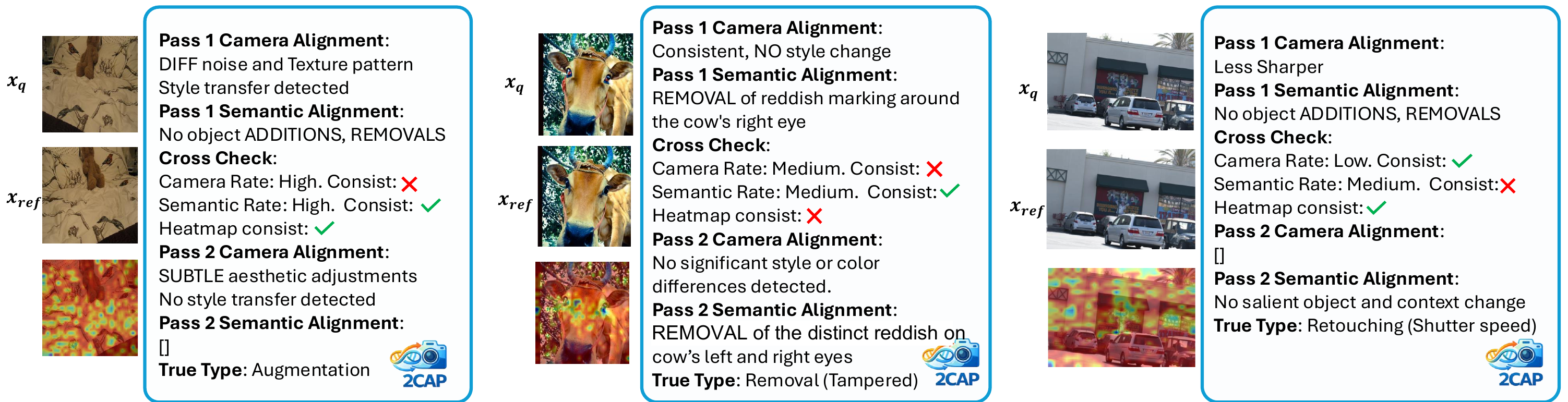}
    \caption{\textbf{\method{} evidence generation case study.} For each query-reference pair, 2CAP performs camera and semantic alignment (Pass 1), applies a reliability cross-check (camera-, semantic-, and heatmap consistency), and optionally uses Pass 2 to finalize the edit type, separating augmentation, tampered removal, and retouching.}
    \label{fig:qualitative-visualization-main}
\end{figure}

This section evaluates the \textit{second} application: \textit{what is the forensic nature of the change?} 
We provide three qualitative examples of \method{} (left–right) in Fig.~\ref{fig:qualitative-visualization-main}. The examples illustrate how alignment cues and saliency inform the cross-check and subsequent revision. More qualitative examples can be found in Appendix~\ref{app:qual_extended}.

\noindent\textbf{Channel score space as VLM context.} Fig.~\ref{fig:score-scatter} (right) visualizes the ($\scam$, $\ssem$) score distribution that we supply as privileged context to the VLM. These scores are raw cosine similarities computed from the pretrained encoders alone, with no fusion module or learned weights involved, and so are independent of the classifier evaluated in Sec.~\ref{sec:four_class}. Retouching clusters tightly in the low-$\scam$, high-$\ssem$ region (camera traces destroyed, semantics preserved), while camera-captured images (including fingerprint-preserved augmentations) cluster at high values on both axes. Tampered images show higher within-class variance because localized edits range from minor inpainting (preserving most camera traces) to large-region replacement (disrupting both channels). The two-dimensional space therefore provides meaningful separation that single-score methods cannot achieve, motivating its use as a feedback signal in the OGR loop. Notably, this category-wise channel separation mirrors the pattern revealed by the trained reliability weights in Fig.~\ref{fig:score-scatter} (left).

We evaluate whether dual-channel scores and saliency heatmaps improve VLM-generated explanations for answering ``what changed.'' We compare VLM baselines with and without our privileged-context feedback. Recent VLM-based approaches (FakeVLM~\cite{wen2025spot}, SIDA~\cite{huang2025sida}, ForgeryGPT~\cite{liu2024forgerygpt})
   address a related but distinct task by fine-tuning VLMs on extensively annotated images with localization and explanations. Direct
   numerical comparison is therefore not straightforward. Instead, our method is plug-and-play and can be combined with these VLM models. To make a reliable comparison, we choose to leverage frozen frontier VLMs with and without our privileged-context feedback.

\noindent\textbf{Setup and metrics.} We evaluate 400 query-reference pairs: 102 benign camera-derived augmentations, 100 retouched pairs, and 198 tampered pairs. For each of four frozen VLMs: GPT-4o-mini, GPT-4o~\citep{achiam2023gpt}, Gemini~2.5 Flash, and Gemini~2.5 Pro~\citep{comanici2025gemini}--the baseline receives the image pair without forensic feedback; +OGR adds dual-channel similarity scores and a saliency heatmap through the Observe-Generate-Refine loop. Both conditions use the same structured output schema. Table~\ref{tab:ogr-main-results} reports change-type accuracy (ChgType), changed-object Recall, action-compatible object Hit, localization proxy IoU, policy-review accuracy/precision/recall, and revision frequency (Rev). Object metrics are evaluated on tampered pairs, and localization uses coarse annotation-derived targets rather than pixel-level masks. Rev measures how often the assessment changes, not the fraction of errors corrected. Metric definitions and policy rules are in Appendices~\ref{app:vlm_eval_details} and~\ref{app:policy}.

\begin{table}[t]
\centering\footnotesize
\newcommand{\ogrstat}[2]{\ensuremath{#1}\,{\scriptsize\color{black!65}\ensuremath{\mathord{\pm}#2}}}
\setlength{\tabcolsep}{2.5pt}
\renewcommand{\arraystretch}{1.12}
\caption{\textbf{VLM evidence generation and downstream policy review.} Baseline versus +OGR on 400 pairs. All values are percentages. Forensic evidence and Rev report mean$\pm$standard deviation over three runs; policy columns retain separate point estimates. Bold marks the better result within each model; $\uparrow$ indicates higher is better. IoU is a localization proxy. Policy metrics use reject as the positive class. Rev is revision frequency and has no preferred direction.}
\label{tab:ogr-main-results}
\begin{tabular*}{\linewidth}{@{\extracolsep{\fill}}llcccccccc@{}}
\toprule
& & \multicolumn{4}{c}{\textbf{Forensic evidence}} & \multicolumn{3}{c}{\textbf{Policy review}} & \\
\cmidrule(lr){3-6}\cmidrule(lr){7-9}
\textbf{Model} & \textbf{Config.} & ChgType $\uparrow$ & Recall $\uparrow$ & Hit $\uparrow$ & IoU $\uparrow$ & Acc. $\uparrow$ & Prec. $\uparrow$ & Recall $\uparrow$ & Rev (\%) \\
\midrule
GPT-4o-mini & Baseline & \ogrstat{66.1}{0.5} & \ogrstat{21.3}{0.9} & \ogrstat{32.5}{1.1} & \ogrstat{25.6}{0.6} & 95.5 & \textbf{94.7} & 51.4 & -- \\
 & +OGR & \ogrstat{\mathbf{72.1}}{0.5} & \ogrstat{\mathbf{29.3}}{0.6} & \ogrstat{\mathbf{39.7}}{1.6} & \ogrstat{\mathbf{36.1}}{1.5} & \textbf{96.2} & 83.3 & \textbf{71.4} & \ogrstat{76.4}{0.3} \\
\midrule
GPT-4o & Baseline & \ogrstat{84.4}{0.3} & \ogrstat{35.6}{0.7} & \ogrstat{59.4}{1.2} & \ogrstat{60.9}{3.7} & 97.8 & 86.1 & 88.6 & -- \\
 & +OGR & \ogrstat{\mathbf{89.8}}{0.2} & \ogrstat{\mathbf{36.4}}{0.2} & \ogrstat{\mathbf{60.3}}{0.3} & \ogrstat{\mathbf{61.3}}{0.3} & \textbf{98.2} & \textbf{86.8} & \textbf{94.3} & \ogrstat{15.2}{0.8} \\
\midrule
Gemini 2.5 Flash & Baseline & \ogrstat{71.0}{1.0} & \ogrstat{41.8}{2.1} & \ogrstat{64.0}{2.3} & \ogrstat{57.4}{1.4} & 92.8 & 56.8 & 71.4 & -- \\
 & +OGR & \ogrstat{\mathbf{78.7}}{2.6} & \ogrstat{\mathbf{46.8}}{0.5} & \ogrstat{\mathbf{72.4}}{0.3} & \ogrstat{\mathbf{70.0}}{1.7} & \textbf{95.2} & \textbf{70.0} & \textbf{80.0} & \ogrstat{36.8}{1.1} \\
\midrule
Gemini 2.5 Pro & Baseline & \ogrstat{73.1}{0.8} & \ogrstat{54.0}{0.2} & \ogrstat{77.6}{1.2} & \ogrstat{73.5}{0.8} & 93.5 & 58.5 & 88.6 & -- \\
 & +OGR & \ogrstat{\mathbf{77.8}}{0.6} & \ogrstat{\mathbf{55.3}}{0.3} & \ogrstat{\mathbf{79.0}}{0.3} & \ogrstat{\mathbf{76.1}}{0.0} & \textbf{94.5} & \textbf{62.7} & \textbf{91.4} & \ogrstat{16.8}{1.7} \\
\bottomrule
\end{tabular*}
\end{table}

\noindent\textbf{Overall results.} OGR improves ChgType, object Recall and Hit, and localization proxy IoU for all four models in Table~\ref{tab:ogr-main-results}. The largest ChgType gains occur for GPT-4o-mini (+6.0 percentage points) and Gemini~2.5 Flash (+7.7 points), showing that the feedback benefits the smaller Mini and Flash variants. GPT-4o and Gemini~2.5 Pro show smaller gains (+5.4 and +4.7 points) and lower revision rates (15.2\% and 16.8\%, versus 76.4\% for Mini and 36.8\% for Flash). This association is consistent with different refinement behavior across models. Given that the grounding annotation for IoU is a proxy, we performed a separate mask-derived-box diagnostic in Appendix~\ref{app:proxy_gt}. Appendix~\ref{app:ogr_ablation} reports the component-ablation study to evaluate the contribution of each component.

\noindent\textbf{Policy review.} We pass the structured VLM evidence to the same rule-based checker in both conditions. Under a predefined mock policy, the checker rejects biometric modifications or additions, removals, and replacements involving people or sensitive entities; other edits are accepted when neither rule triggers. It uses regular-expression and lexicon matching, with safe-phrase handling in the biometric branch, rather than assigning verdicts directly from channel scores. We compare pair-level decisions with 400 human-annotated verdicts (365 accept, 35 reject). Because an always-accept rule reaches 91.25\% accuracy, Table~\ref{tab:ogr-main-results} also reports precision and recall with reject as the positive class. OGR improves accuracy and rejection recall for every model. GPT-4o has the highest +OGR accuracy (98.2\%) and rejection recall (94.3\%). For GPT-4o-mini, rejection recall increases from 51.4\% to 71.4\%, while precision decreases from 94.7\% to 83.3\%; the other models improve both. These results concern a fixed mock policy and the combined evidence-and-checker pipeline, rather than general policy compliance. Appendix~\ref{app:policy} details the rules, input fields, and verdict aggregation.

\section{Conclusion}
\label{sec:conclusion}
We presented \method{}, combining camera-related and semantic representations for reference-free authenticity classification and reference-based evidence generation. Channel ablations support combining both representations on the evaluated benchmark. The learned gate distributions vary by category, but do not establish causal attribution to the original channels; the matched single-seed comparison does not demonstrate a performance benefit from reliability gating. For reference-based assessment, dual-channel similarity scores and saliency maps guide frozen VLMs through OGR without forensic instruction tuning. Across four evaluated VLM configurations, full OGR improves change-type accuracy over the single-pass baseline, with model-dependent trade-offs in the component ablations. The mock-policy evaluation illustrates how structured evidence can support a downstream decision under explicit rules. These results motivate distinguishing \textit{what changed} from \textit{whether it matters under a given policy}.

\bibliographystyle{unsrtnat}
\bibliography{references}

\newpage
\section*{Appendix}
\appendix



\section{Dataset Details}
\label{sec:appendix_dataset}
We curate a multi-source dataset comprising four categories: camera-captured images, fully synthetic images, tampered images, and GenAI-retouched images. This section provides full details on data composition, construction methodology, and the measures taken to ensure evaluation integrity.

To prevent data leakage, we enforce strict image-level disjointness: no image (camera-captured or edited) appearing in any evaluation split is used during either Stage~A (contrastive pretraining) or Stage~B (fusion training). For datasets that contribute to both training and evaluation (e.g., SID-Set~\citep{huang2025sida}, Defactify~\citep{roy2026comprehensive}), we use the official train/test splits. For held-out datasets (VISION~\citep{shullani2017vision}, Flickr8K~\citep{huang2025ru}), zero training images are drawn from these sources. We check exact-image duplication using image hashes. This check does not establish that an evaluation image's original photograph, crop, or other edited version is absent from training; the parent-image audit described in Appendix~\ref{app:source_control_plan} is an additional requirement.

\subsection{Training Data for Contrastive Learning}
We focus on datasets whose real images are verifiably camera-captured photographs, excluding common image-editing benchmarks (\textit{e.g.}, ImgEdit~\citep{ye2025imgedit}) that contain specialized content such as remote sensing or document images. Such non-photographic imagery would weaken the camera-fingerprint signal our encoder aims to learn.

Table~\ref{tab:stage1_data} details the training data composition for contrastive learning.

\begin{table}[htbp]
\centering
\caption{\textbf{Training data for contrastive learning.} Paired sources provide camera-edit correspondences; unpaired sources contribute additional camera and edited images. The total counts image entries ($2\times\StageAPairs+\StageADresden+\StageASIDCamera+\StageAUnpairedEdited$), not deduplicated unique images. The unique-image count requires a per-image manifest.}
\label{tab:stage1_data}
\begin{tabularx}{\linewidth}{Xlr}
\toprule
\textbf{Source} & \textbf{Type} & \textbf{Size} \\
\midrule
MagicBrush~\citep{zhang2023magicbrush} + SAGI-D~\citep{giakoumoglou2025sagi} & Paired (camera-edit) & \makecell{\StageAPairs \\ pairs} \\
Dresden~\citep{gloe2010dresden} & Unpaired camera & \StageADresden \\
SID-Set~\citep{huang2025sida} & Unpaired camera & \StageASIDCamera \\
SID-Set~\citep{huang2025sida} & Unpaired edited & \StageAUnpairedEdited \\
\midrule
\textbf{Total image entries} & & \textbf{\StageAEntries} \\
\bottomrule
\end{tabularx}
\end{table}

\subsection{Evaluation Data Statistics}

Table~\ref{tab:dataset_sizes} summarizes the historical downstream training pool and source-level evaluation inventory; the controlled four-class split is described in Sec.~\ref{sec:four_class}.

\begin{table}[htbp]
\centering
\caption{\textbf{Historical downstream training pool and source-level evaluation inventory.} The historical training pool contains 12{,}800 images (3{,}200 per class); the controlled fusion experiments instead use the filtered 9{,}877/2{,}451 training/validation split described in Sec.~\ref{sec:four_class}. The common four-class test split in Table~\ref{tab:four_class_results} samples 800 images per class from the evaluation pools, for 3{,}200 images total; the inventory counts below describe the larger pools rather than the size of that test split. * indicates not seen in contrastive learning.}
\label{tab:dataset_sizes}
\begin{tabular}{lrr}
\toprule
\textbf{Dataset} & \textbf{Historical training pool} & \textbf{Evaluation inventory} \\
\midrule
{SAGI-D}~\citep{giakoumoglou2025sagi}                      & 2{,}240 & 3{,}838 \\
{Defactify*}~\citep{roy2026comprehensive}                  & 3{,}840 & 22{,}000 \\
{SID-Set}~\citep{huang2025sida}                           & 640    & 16{,}000 \\
{MagicBrush}~\citep{zhang2023magicbrush}                 & 2{,}240 & 0 \\
{Dresden}~\citep{gloe2010dresden}                         & 640    & 0 \\
{Generative Photography*}~\citep{yuan2025generative}       & 3{,}200 & 800 \\
{Flickr8K*} (\textit{held-out})~\citep{huang2025ru}        & 0      & 3{,}156 \\
{VISION*} (\textit{held-out})~\citep{shullani2017vision}   & 0      & 6{,}000 \\
\bottomrule
\end{tabular}
\end{table}



\subsection{Dataset Construction Details}
\label{sec:appendix_dataset_details}

\noindent\textbf{Camera images.}
We aggregate authentic photographs from complementary sources. To capture classic device-forensics conditions, we include images from the Dresden Image Database (Dresden) \citep{gloe2010dresden}, which is widely used for camera model identification, and VISION \citep{shullani2017vision}, which includes native captures as well as versions processed by online sharing pipelines. To better reflect modern content distributions encountered in real-vs-synthetic detection, we additionally use the authentic splits of Defactify \citep{roy2026comprehensive} and SID-Set \citep{huang2025sida}. Paired camera originals from SAGI-D and MagicBrush additionally contribute to contrastive pretraining (Table~\ref{tab:stage1_data}); the source examples in Appendix~\ref{app:exp-details} are illustrative rather than an exhaustive depiction of any one split.

\noindent\textbf{Fully synthetic images.}
We define \textit{full synthesis} as images generated entirely by GenAI from text prompts, with no camera capture involved. These images lack any authentic camera fingerprints and may exhibit subtle semantic inconsistencies (lighting, geometry, physics). We compile images from diverse text-to-image generators: Defactify~\citep{roy2026comprehensive} (multiple popular models including SD2.1, SDXL, SD3, DALL-E~3, Midjourney~6) and Flickr8K-derived generations~\citep{huang2025ru} (controlled text-conditioned synthesis). The absence of camera capture does not determine either learned channel response: synthesis can be semantically coherent, and learned camera-related features can respond to processing artifacts. In particular, reference-based cosine scores depend on the chosen query--reference pair and are not fixed by the synthesis label.

\noindent\textbf{Tampered images.}
We define \textit{tampering} as localized GenAI editing that modifies semantic content in specific image regions while potentially preserving camera traces elsewhere. This includes object insertion, removal, replacement, and inpainting operations. We include three complementary families:
SID-Set~\citep{huang2025sida} (social-media motivated edits), SAGI-D~\citep{giakoumoglou2025sagi} (AI-based inpainting and region-level modifications), and MagicBrush~\citep{zhang2023magicbrush} (instruction-guided editing triplets). Tampering can change semantic content locally while leaving camera-related cues elsewhere; small or coherent edits need not produce low global semantic similarity. Channel responses can overlap with those of other modification types.

\noindent\textbf{GenAI-retouched images.}
We define \textit{retouching} here as GenAI-enabled camera-parameter restyling intended to preserve scene content. We use the Generative Photography dataset~\citep{yuan2025generative}, covering bokeh, color temperature, focal length, and shutter speed. The content-preservation criteria and limits of this dataset are discussed in Appendix~\ref{app:retouch_scope}.

\subsection{Retouching Dataset Scope}
\label{app:retouch_scope}

The retouching evaluation uses GP examples covering bokeh, color temperature, focal length, and shutter speed (motion blur). The intended operation changes photographic appearance while preserving scene content; this intention does not guarantee that every generated output meets the definition. Object identity, object presence, and scene layout must be checked independently of detector predictions. Photometric changes affecting foreground or background can be retouching; replacing foreground or background content cannot. Focal-length changes that reveal or remove content at crop boundaries require an explicit annotation policy.

The current experiments do not establish transfer to an independent GenAI retouching generator. Different effects or checkpoints within GP are not independent generator families. The classical OpenCV diagnostic is informative about processing sensitivity but does not close this gap. Broader portrait, computational-photography, and mixed-operation settings also remain outside the validated scope. Suitable external data would require genuine camera originals, documented generative editing provenance, and verified content preservation. Availability of a general image-editing model or a physically simulated retouching target set is not sufficient by itself.

\subsection{Proposed Controls for Source and Processing Confounds}
\label{app:source_control_plan}
\noindent\textbf{Status.} The following controls are proposed additional analyses; no results from them are reported in this paper. They are designed to use existing images where provenance and original-edit links can be recovered. The within-source, OpenCV, and mild-processing results in Appendices~\ref{app:within_source}--\ref{app:benign_diagnostic} are separate, already reported diagnostics.

\paragraph{Parent-image separation.}
Create a joint manifest for Stage~A, Stage~B, validation, and evaluation. For every image, record the source collection, original-photo identifier, derivative links, operation, tool/checkpoint, and split. Group each original with all crops, recompressions, and edited descendants before splitting. Check both provenance identifiers and exact/near-duplicate image matches across datasets; related source collections can contain the same original under different identifiers. All evaluation groups must be absent from training and model selection. If overlap is found, retrain from the earliest affected stage and rerun the affected comparisons. Report this task-specific audit separately from unknown foundation-model pretraining exposure.

\paragraph{Matched-original comparison.}
Where genuine camera originals for GP outputs are available, evaluate the original and its retouched version using the same frozen checkpoint, restricted to original-photo groups excluded from both training stages. Report camera recall, retouched recall, the four-way confusion matrix, and the paired change in retouched probability. Preserve all four output classes; do not silently discard predictions assigned to tampered or synthetic. This comparison reduces content/source differences between the two classes, but remains a within-GP evaluation. If original-camera provenance cannot be recovered, this control cannot be claimed.

\paragraph{Processing controls.}
Evaluate native files and a separately reported common-export condition. Apply the same specified resizing and encoding policy to originals and derivatives, and record original dimensions, codec, and export settings. Report the change in performance rather than treating normalization as guaranteed to preserve forensic traces. Use the existing mild-processing and OpenCV examples only after resolving parent overlap and documenting operation parameters and labels. Under a GenAI-specific definition, conventional processing is not positive evidence of GenAI provenance; examine retouched false alarms separately from all non-camera predictions. A source/format-only diagnostic can additionally test how predictive these nuisance variables are of the labels.

\paragraph{Leave-one-effect-out evaluation.}
If the existing GP samples have reliable effect labels, construct four folds: exclude bokeh, color temperature, focal length, or shutter-speed examples in turn from all task-specific training and validation, while maintaining parent-image separation. Retrain the affected components and test the excluded effect on unseen original-photo groups. Ensure that another derivative of a test original does not appear among the three training effects. Keep the remaining-class evaluation protocol fixed and report each fold separately, alongside a control trained with all four effects using a matched training budget. This tests generalization to a held-out photographic effect within GP; it is not a test of an unseen generator.

\paragraph{Comparison and uncertainty.}
Use the same evaluation groups for full \method{}, semantic-only and camera-only variants, and applicable baselines. Report effect- and source-specific results, sample counts in original-photo groups as well as derivative images, and confidence intervals resampled at the original-photo-group level. Select thresholds and processing settings on validation data. Freeze these choices before testing, and distinguish training-seed variability from uncertainty over photographs.

\paragraph{Remaining external-data requirement.}
None of these controls removes the need for an independent GenAI retouching pipeline if claiming cross-generator transfer. Such a test requires camera-original/generated-retouch pairs with documented provenance and content preservation. An editor that replaces scene content, a set of conventional simulated targets, or a new checkpoint within the same family does not automatically meet that requirement. Until suitable data become available, the retouching claim remains limited to GP and the reported processing diagnostics.

\subsection{Metric Choice Rationale}
\label{app:metrics}

We select metrics to match the evaluation goal of each experiment:

\paragraph{Camera authenticity detection.}
We report AUC, precision, recall, and F1 for the four-class classification task in Table~\ref{tab:four_class_results}. Each image receives a single predicted label using four-way argmax, $\hat{y}=\arg\max_{k\in\{1,2,3,4\}}p_k$. Per-class AUC is computed one-vs-rest from the class scores; per-class precision, recall, and F1 use the argmax predictions. Overall AUC, precision, recall, and F1 are macro averages: the unweighted arithmetic mean of the corresponding four per-class values.

\paragraph{Cross-generator generalization.}
We report \textbf{AUC only}, as the goal is to assess ranking performance across diverse generators rather than to fix a specific operating point. Per-generator AUC isolates how well each method generalizes to individual generator families, while the \textbf{Average} column summarizes overall robustness.

\paragraph{Held-out camera-only datasets (VISION).}
For datasets containing only authentic camera images (a single class), we report the \textbf{Brier score}:
\[
\text{Brier Score} = \frac{1}{N} \sum_{i=1}^{N} (p_i - y_i)^2,
\]
where $p_i$ is the predicted camera probability and $y_i=1$ for every VISION image. This camera-only score measures squared error on authentic images, but cannot by itself establish calibration across authentic and manipulated classes. AUC requires both positive and negative classes and is therefore not applicable to single-class datasets.

\paragraph{VLM evidence generation (Sec.~\ref{sec:vlm_eval}).}
We evaluate 400 query--reference pairs (198 tampered, 100 retouched, 102 augmented), comparing the visual-only baseline with full OGR and its component ablations. Table~\ref{tab:ogr-main-results} reports the main evidence-generation and policy-review results. Table~\ref{tab:ogr-ablation} extends the baseline/+OGR comparison with component ablations and reports change-type accuracy, object Recall and Hit, localization proxy IoU, and revision rate as means with run standard deviations over three runs for the component-ablation study. Detailed metric definitions appear in Appendix~\ref{app:vlm_eval_details}.

\section{Extended Related Work}
\label{app:relatedwork}

Several concurrent works address related problems. We discuss
key differences and clarify \method{}'s positioning.

\paragraph{Classical forensic cues.}
Earlier image forensics focused on hand-crafted traces left by image
formation and post-processing pipelines. Sensor-pattern-noise methods
exploit device-specific PRNU fingerprints for source attribution and
integrity analysis~\citep{lukas2006digital,chen2008determining}, while
JPEG-based analysis detects recompression inconsistencies and local block
artifacts introduced by splicing or resaving~\citep{bianchi2012image}.
Related statistical approaches also localize tampered regions from
distributional irregularities in transform-domain statistics, such as
first-digit features~\citep{amerini2014splicing}. These approaches are
physically grounded and often interpretable, but their effectiveness can
degrade under resizing, heavy post-processing, or when manipulations do
not leave stable low-level traces. In contrast, \method{} retains access
to low-level forensic evidence while embedding it into a learned
reference-based comparison framework.

\paragraph{Learning-based image forgery detection and localization.}
Deep IFDL methods replace hand-crafted cues with end-to-end feature
learning for tamper detection and localization. ManTra-Net~\citep{wu2019mantra}
models local anomaly patterns across a wide range of manipulations,
MVSS-Net~\citep{dong2022mvss} combines multiview and multiscale
supervision for manipulation segmentation, and TruFor~\citep{guillaro2023trufor}
integrates complementary clues including RGB content and forensic
residuals to improve trustworthy localization, building on learned camera
fingerprints such as Noiseprint~\citep{cozzolino2020noiseprint}. These
methods have substantially improved robustness over classical pipelines,
but they remain primarily single-image predictors trained for fixed
output spaces (labels or masks). \method{} is complementary: rather than
predicting authenticity from a single image alone, it explicitly compares
a query against a reference and returns evidence in a metric space that
supports verification, retrieval, and downstream reasoning.

\paragraph{Early AI-generated image detection.}
A parallel line of work studies artifacts specific to synthetic image
generation. Early methods analyzed spectral and checkerboard artifacts in
GAN outputs~\citep{zhang2019detecting,frank2020leveraging}, while others
showed that generative models leave characteristic fingerprints that can
support both detection and attribution~\citep{marra2019gans,yu2019attributing}.
CNNSpot~\citep{wang2020cnn} demonstrated that CNN-generated images were
surprisingly easy to distinguish under in-domain conditions, helping
establish cross-model generalization as a central evaluation criterion.
This literature highlights the importance of low-level generator traces,
but most methods still formulate the task as direct classification. By
contrast, \method{} uses low-level forensic cues as one component of a
reference-based embedding, which better supports fine-grained comparison
across images and manipulation settings.

\paragraph{Generalizable detection across generators.}
More recent work emphasizes robustness to unseen synthesis models and
post-processing. UnivFD~\citep{ojha2023towards} shows that pretrained
representations can provide strong transfer across diverse generators with
minimal task-specific adaptation, while DIRE~\citep{wang2023dire} exploits
reconstruction discrepancies induced by diffusion inversion for
diffusion-image detection. Frequency-oriented methods continue this trend
by targeting generator-dependent spectral artifacts, including
AIDE~\citep{yan2024sanity}. D3~\citep{yang2025d} further scales
multi-generator discrepancy learning with shallow detection heads over
pretrained features. Relative to these approaches, \method{} is less tied
to a fixed discriminative decision rule: its contrastive objective and
reference-conditioned inference allow the same representation to support
verification, nearest-neighbor evidence retrieval, and compatibility with
frozen VLM-based reasoning modules.

\paragraph{Specialized manipulation domains.}
Some related work focuses on narrower but practically important
manipulation families, especially facial retouching. Earlier methods
study supervised retouching detection from facial appearance
changes~\citep{bharati2016detecting,bharati2017demography}, and more
recent datasets such as RetouchingFFHQ~\citep{ying2023retouchingffhq}
enable fine-grained benchmarking at larger scale. Diffusion-based
restoration or reversal has also been explored for authenticity analysis
in this setting~\citep{frrffusion2024}. These methods are valuable for
domain-specific benchmarks, but they are typically tailored to a
particular manipulation family. \method{} instead targets a broader
verification setting spanning AI-generated imagery and image-level
forensic comparison without assuming a single manipulation type.

\paragraph{Model provenance and metadata-based authenticity.}
Content provenance standards such as C2PA~\citep{c2pa2024} provide a
complementary route to authenticity by attaching cryptographically
verifiable provenance metadata at creation or editing time. Such
approaches are important for deployment ecosystems, but they depend on
adoption across capture and editing pipelines and may be unavailable for
legacy or redistributed content. \method{} is orthogonal to provenance
standards: it operates directly on image evidence and remains applicable
when trusted metadata is absent, stripped, or incomplete.

\paragraph{Positioning of \method{}.}
The literature spans (i) classical low-level forensic
analysis, (ii) end-to-end image forgery detection and localization,
(iii) generalizable AI-generated image detection, and (iv) fine-tuned
VLM-based explainable detectors. \method{} occupies a different point in
this design space. We decompose GenAI manipulations along two complementary
information channels: a \textit{camera channel} that encodes sensor-level
traces, and a \textit{semantic channel} that encodes scene-level coherence.
On top of this shared dual-channel foundation we build two applications.
\textbf{(1)~Authenticity classification:} a four-class classifier that
distinguishes camera-captured, fully synthetic, tampered, and
GenAI-retouched images via cross-attention fusion with learnable
reliability weights, moving beyond the binary real/fake formulation that
dominates prior work in (i)-(iii). \textbf{(2)~VLM evidence generation:}
dual-channel alignment scores and patch-level saliency are supplied to a
\textit{frozen} VLM through an Observe-Generate-Refine loop, eliminating
forensic instruction tuning used by
FakeVLM~\citep{wen2025spot}, SIDA~\citep{huang2025sida}, and
ForgeryGPT~\citep{liu2024forgerygpt} in (iv). The dual-channel
decomposition is the unifying primitive that lets a single framework
address both authenticity classification and interpretable evidence
generation, while the frozen-VLM design avoids the annotation and
retraining costs of supervised forensic VLM pipelines.

\section{Experimental Details}
\label{app:exp-details}

\subsection{Visualization of Data}
\label{app:data_visualization}

We visualize source-dataset examples from the four modification categories. These illustrations show the available source material and do not define the composition of the common test split.

\begin{figure}[h]
\centering
\includegraphics[width=\linewidth]{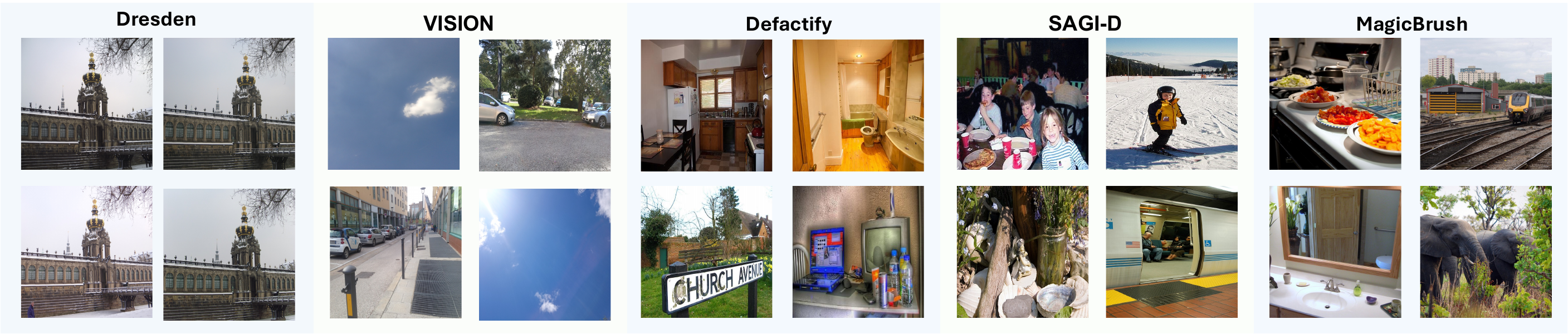}
\caption{\textbf{Visualization of camera-captured images} from the source collections shown. For Dresden and VISION, we show images from diverse camera models (Agfa, Canon, Nikon, Samsung, Huawei, iPhone, Sony, etc). The source names identify the illustrated collections; split membership is specified separately in the evaluation protocol.}
\label{fig:vis_camera}
\end{figure}

\paragraph{Camera images.}
Figure~\ref{fig:vis_camera} shows authentic camera-captured photographs from diverse sources (Dresden, VISION, Defactify, SAGI-D, MagicBrush). These span a wide range of scenes, lighting conditions, and camera models.

\begin{figure}[h]
\centering
\includegraphics[width=\linewidth]{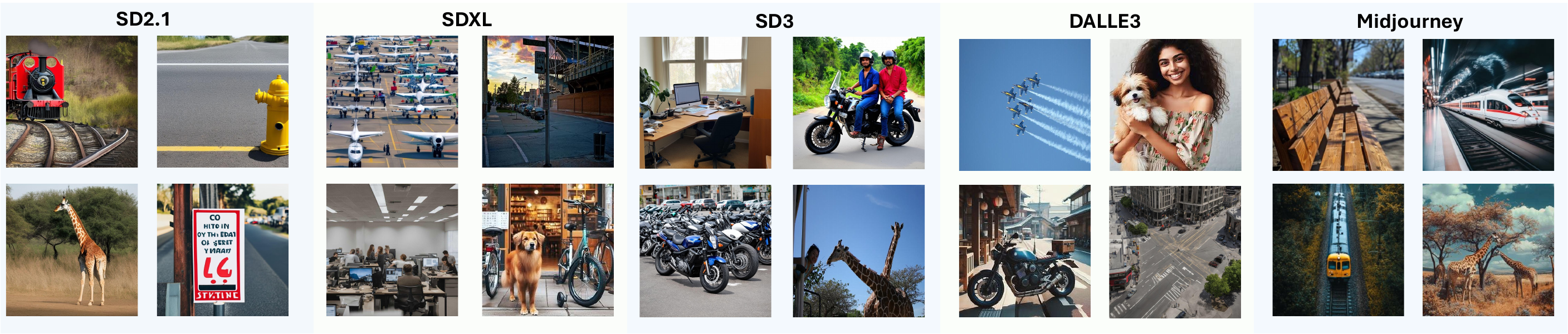}
\caption{\textbf{Visualization of fully synthetic images} generated by different text-to-image models. These images contain no camera traces and may show subtle semantic inconsistencies in lighting, geometry, or physics.}
\label{fig:vis_synthetic}
\end{figure}

\paragraph{Fully synthetic images.}
Figure~\ref{fig:vis_synthetic} shows text-to-image generations from multiple generators (SD2.1, SDXL, SD3, DALL-E~3, Midjourney~6). These images lack camera fingerprints entirely and may exhibit subtle semantic artifacts.

\begin{figure}[h]
\centering
\includegraphics[width=\linewidth]{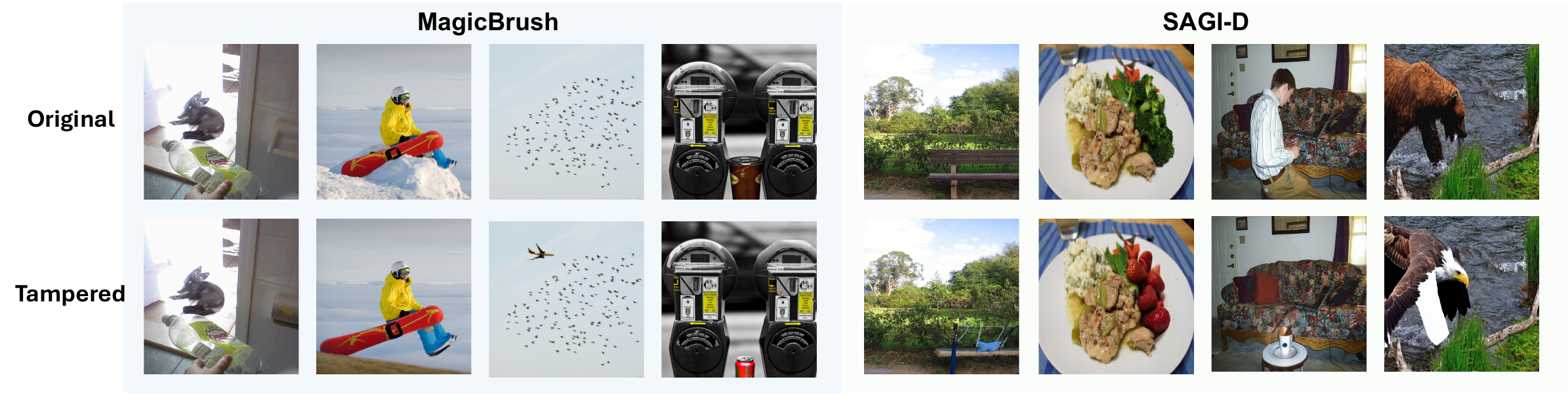}
\caption{\textbf{Visualization of tampered images} with their camera-captured references. Tampered images contain localized semantic edits; their effects on semantic similarity and camera-related cues depend on the edit and can overlap with other categories.}
\label{fig:vis_tampered}
\end{figure}

\paragraph{Tampered images.}
Figure~\ref{fig:vis_tampered} shows localized GenAI edits (object insertion, removal, replacement, inpainting) from SAGI-D and MagicBrush. Each example is shown alongside its camera-captured reference to highlight the semantic modifications.

\begin{figure}[h]
\centering
\includegraphics[width=\linewidth]{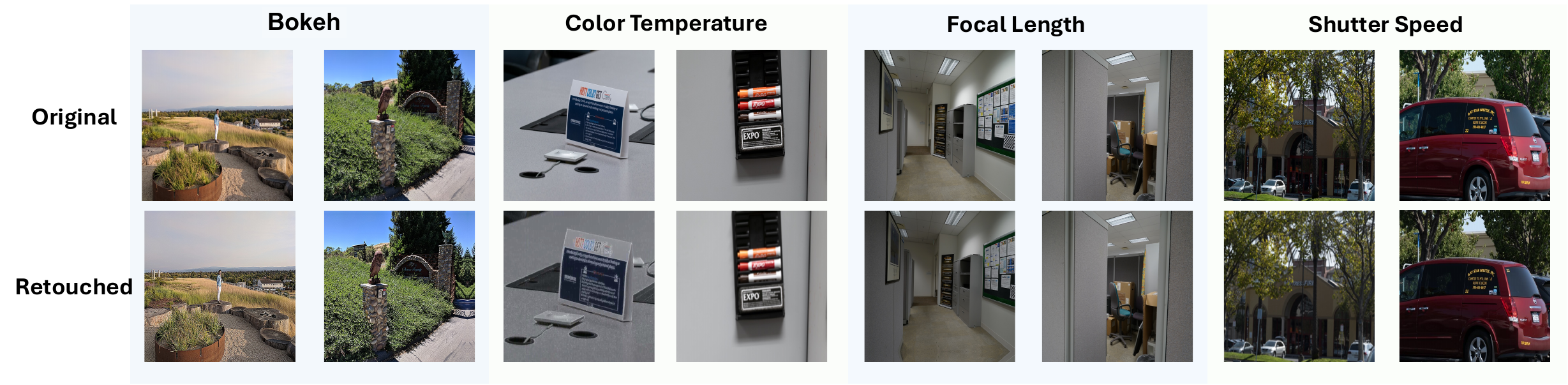}
\caption{\textbf{Visualization of GenAI-retouched images} with their original camera captures. These examples show retouching with low camera-feature similarity and high semantic similarity; actual responses may overlap with tampering.}
\label{fig:vis_retouched}
\end{figure}

\paragraph{GenAI-retouched images.}
Figure~\ref{fig:vis_retouched} shows global camera-parameter restyling from the Generative Photography dataset, covering bokeh adjustment, color temperature shift, focal length change, and shutter speed modification. Each example is paired with its original camera capture.

\subsection{Model Architecture and Training Details}
\label{app:architecture}

\begin{figure}[h]
    \centering
    \includegraphics[width=0.9\textwidth]{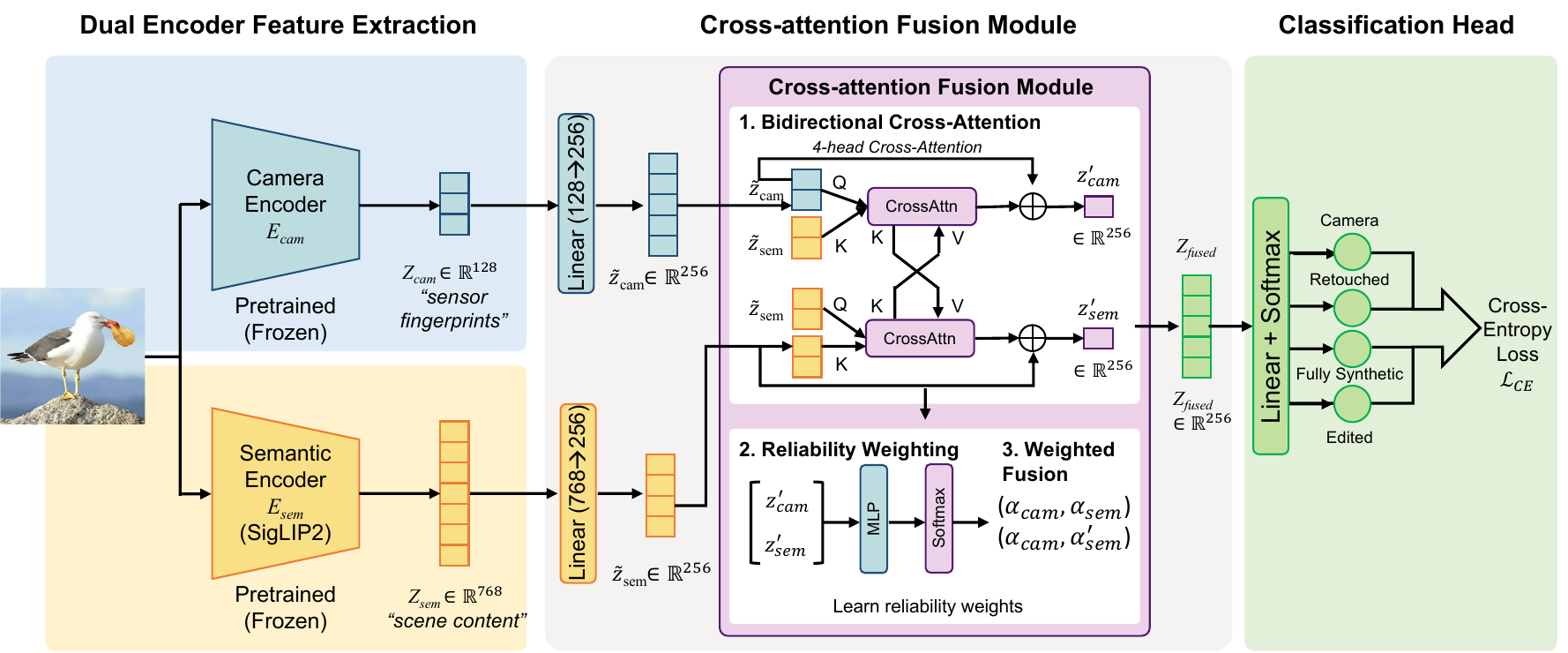}
    \caption{\textbf{Cross-channel attention fusion for authenticity classification.} Camera features $\mathbf{z}_{\mathrm{cam}}$ and semantic features $\mathbf{z}_{\mathrm{sem}}$ are projected to a shared 256-d space, then refined via bidirectional 4-head cross-attention. Learnable reliability weights $(\alpha_{\mathrm{cam}}, \alpha_{\mathrm{sem}})$ adaptively combine the channel representations based on input characteristics. The fused representation is classified as camera-captured or edited.}
    \label{fig:cross-pip}
\end{figure}

We provide complete implementation details for reproducibility. The 2CAP classification pipeline is depicted in Figure.~\ref{fig:cross-pip}

\noindent\textbf{Camera encoder $\Ecam$.}
We use a DCT-domain dual-branch pipeline with patch-level frequency analysis~\citep{ahmed1974discrete}. Images are split into $32 \times 32$ patches, ranked by DCT frequency energy, and selected patches are processed with 30 SRM steganalysis filters~\citep{yan2024sanity} to extract noise residuals. Two ResNet-50 backbones~\citep{he2016deep} separately encode high-frequency (textured) and low-frequency (smooth) patches; branch features are fused by element-wise addition and projected to 128-d unit-norm embeddings.

\noindent\textbf{Semantic encoder $\Esem$.}
We use frozen SigLIP2~\citep{tschannen2025siglip2} (ViT-base) to extract 768-d global semantic embeddings. For the saliency map, we extract 27$\times$27 patch-level features and compute pairwise cosine similarities between query and reference patches.

\noindent\textbf{Cross-attention fusion.}
The camera and semantic patch-token sequences retain their respective patch counts when projected to 256-d via learned linear layers (Sec.~\ref{sec:fusion}). Bidirectional 4-head cross-attention (2 layers) normalizes over key patches in each direction. After channel-wise pooling, a 2-layer gating MLP takes the concatenated vectors and produces two logits; a softmax over channels converts these into input-dependent weights that sum to one. A 2-layer classification head (256$\rightarrow$128$\rightarrow$4) outputs probabilities over camera-captured, retouched, tampered, and fully synthetic images. Table~\ref{tab:cross-architecture} lists the tensor shapes and normalization axes. The two patch counts need not be equal: each head forms an $N_{\mathrm{cam}}\times N_{\mathrm{sem}}$ attention matrix in one direction and its reverse shape in the other. Pooling is applied only after cross-attention, rather than reducing the inputs to singleton global tokens before attention.

\begin{table}[htbp]
\centering\small
\setlength{\tabcolsep}{5pt}
\renewcommand{\arraystretch}{1.22}
\caption{\textbf{Cross-channel fusion architecture and tensor shapes.} $B$ is batch size; $N_{\mathrm{cam}}$ and $N_{\mathrm{sem}}$ are channel-specific patch counts. Shapes include the batch axis. Attention uses four heads and two layers; the attention-weight shapes below are per head. The key-patch, channel, and class softmax operations normalize over different axes.}
\label{tab:cross-architecture}
\begin{tabularx}{\linewidth}{@{}p{0.18\linewidth}p{0.32\linewidth}X@{}}
\toprule
\textbf{Stage} & \textbf{Tensor / shape} & \textbf{Operation and normalization} \\
\midrule
Patch tokens & $Z_{\mathrm{cam}}: B\times N_{\mathrm{cam}}\times128$\newline
$Z_{\mathrm{sem}}: B\times N_{\mathrm{sem}}\times768$ & Retain the patch sequence in each channel; no global pooling before attention. \\
\addlinespace[3pt]
Linear projections & $\widetilde Z_c: B\times N_c\times256$ & Project camera and semantic features independently to 256 dimensions. \\
\addlinespace[3pt]
Camera-query\newline attention & $A_{\mathrm{cam}}: B\times N_{\mathrm{cam}}\times N_{\mathrm{sem}}$\newline
$Z'_{\mathrm{cam}}: B\times N_{\mathrm{cam}}\times256$ & Camera queries; semantic keys and values. Softmax over the $N_{\mathrm{sem}}$ keys for each query and head. \\
\addlinespace[3pt]
Semantic-query\newline attention & $A_{\mathrm{sem}}: B\times N_{\mathrm{sem}}\times N_{\mathrm{cam}}$\newline
$Z'_{\mathrm{sem}}: B\times N_{\mathrm{sem}}\times256$ & Semantic queries; camera keys and values. Softmax over the $N_{\mathrm{cam}}$ keys for each query and head. \\
\addlinespace[3pt]
Channel pooling & $\mathbf z'_c: B\times256$ & Pool over each channel's patch axis after residual cross-attention. \\
\midrule
Reliability gate & $[\mathbf z'_{\mathrm{cam}};\mathbf z'_{\mathrm{sem}}]: B\times512$\newline
$\boldsymbol\ell,\boldsymbol\alpha: B\times2$ & Two-layer MLP produces two logits. Channel-axis softmax gives per-image weights with $\alpha_{\mathrm{cam}}+\alpha_{\mathrm{sem}}=1$. \\
\addlinespace[3pt]
Weighted fusion & $\mathbf z_{\mathrm{fused}}: B\times256$ & $\alpha_{\mathrm{cam}}\mathbf z'_{\mathrm{cam}}+\alpha_{\mathrm{sem}}\mathbf z'_{\mathrm{sem}}$. \\
\addlinespace[3pt]
Classification & $\mathbf p: B\times4$ & MLP $256\to128\to4$, followed by softmax over four classes; cross-entropy training. \\
\bottomrule
\end{tabularx}
\end{table}

\noindent\textbf{Contrastive pretraining for $\Ecam$.}
We pretrain $\Ecam$ on \StageAEntries{} image entries (not a verified unique-image count) with InfoNCE loss (data composition in Table~\ref{tab:stage1_data}). Positives include mild augmentations (horizontal flip, mild color jitter $\pm$0.1, JPEG quality $\geq$90) and, when available, same-camera images; negatives include paired edits and batch negatives. Training uses AdamW~\citep{loshchilov2019decoupled} optimizer (lr $3\times10^{-4}$, weight decay 0.05, $\beta_1=0.9$, $\beta_2=0.999$) for 20 epochs on 4$\times$ NVIDIA A100 80GB GPUs with batch size 512. Temperature-weighted sampling balances dataset sizes. Augmentations avoid blur and heavy JPEG recompression to limit processing changes; this does not guarantee preservation of sensor fingerprints.

\noindent\textbf{Supervised fusion training.}
With both encoders frozen, we train only the fusion module (cross-attention layers + reliability MLP + classification head, $\sim$2M parameters) on the supervised four-class authenticity task. For the controlled comparison in Sec.~\ref{sec:four_class}, all six variants use the same 9{,}877 training and 2{,}451 validation images after source-reference overlap removal, deduplication, and group-based partitioning. Training uses AdamW (lr $10^{-3}$) with a cosine annealing schedule, batch size 2048, and a 20-epoch budget. Validation macro-F1 selects each checkpoint before a single evaluation on the common 3{,}200-image test split (800 images per class). Cross-entropy loss with label smoothing (0.1) is used.

\subsection{Architectural Comparison with AIDE}
\label{app:aide_diff}

Table~\ref{tab:aide_comparison} explicitly delineates the architectural
differences between \method{}'s camera encoder and AIDE~\citep{yan2024sanity},
which inspired the frequency-aware patch selection design.

\begin{table}[h]
\centering
\caption{\textbf{Architectural comparison: \method{} camera encoder vs.\ AIDE.}}
\label{tab:aide_comparison}
\begin{tabular}{@{}lll@{}}
\toprule
Component & AIDE & \method{} \\
\midrule
Patch selection & DCT energy ranking & DCT energy ranking (same) \\
Filter bank & 30 SRM filters & 30 SRM filters (same) \\
Backbone & Dual-branch ResNet-50 & Dual-branch ResNet-50 (same) \\
Training objective & Cross-entropy (binary) & InfoNCE contrastive \\
Training data & Single-image labels & Paired camera-edit + unpaired \\
Embedding space & Class logits & 128-d unit-norm (metric space) \\
Downstream use & Binary classifier & Cosine similarity $\scam$ \\
Semantic channel & CLIP + linear proj. & Frozen SigLIP2 + cross-attn \\
VLM integration & None & OGR privileged-context loop \\
\bottomrule
\end{tabular}
\end{table}

\paragraph{Summary.} The low-level feature extraction pipeline is shared
with AIDE; \method{}'s novelty lies in (1)~contrastive pretraining for
a metric embedding space enabling reference-based verification,
(2)~dual-channel decomposition with cross-attention fusion, and
(3)~VLM evidence synthesis via OGR.

\subsection{Inference Pipeline}
\label{app:pseudocode}
Stage~A denotes camera-encoder contrastive pretraining and Stage~B denotes supervised fusion training; OGR is the subsequent frozen-VLM inference procedure for Q2. Algorithm~\ref{alg:inference} summarizes the inference procedure of \method{}.
For each channel $c\in\{\mathrm{cam},\mathrm{sem}\}$, $z_c^v=\mathcal E_c^g(\mathbf x_v)$ denotes a global embedding, whereas $Z_c^v=\mathcal E_c^{\mathrm{tok}}(\mathbf x_v)$ is a patch-token sequence, for $v\in\{q,r\}$. The camera global embedding is the 128-d unit-norm contrastive representation. The superscripts $g$ and $\mathrm{tok}$ denote distinct outputs of the same encoder, not additional encoders. Q1 uses only query tokens: cross-attention precedes pooling, gating, and four-class classification. Q2 compares global query/reference embeddings for cosine scores and semantic patch tokens for saliency; these signals guide the frozen VLM. Q1's four-class prediction and Q2's three-class change assessment are distinct outputs; the reference is required only for Q2.

\begin{algorithm}[h]
\caption{\method{} Inference Pipeline}
\label{alg:inference}
\begin{algorithmic}[1]
\REQUIRE Query image $\xq$, reference image $\xref$
\ENSURE Q1 four-class prediction, Q2 forensic explanation, saliency map
\STATE $z_c^v \leftarrow \mathcal E_c^g(\mathbf x_v)$ for $c\in\{\mathrm{cam},\mathrm{sem}\}$, $v\in\{q,r\}$
  \hfill \textit{// Global embeddings for Q2}
\STATE $Z_c^q \leftarrow \mathcal E_c^{\mathrm{tok}}(\xq)$ for $c\in\{\mathrm{cam},\mathrm{sem}\}$
  \hfill \textit{// Query patch tokens for Q1}
\STATE $Z_{\mathrm{sem}}^r \leftarrow \Esem^{\mathrm{tok}}(\xref)$
  \hfill \textit{// Reference patch tokens for Q2}
\STATE $\scam \leftarrow \cos(z_{\mathrm{cam}}^q,z_{\mathrm{cam}}^r)$; $\ssem \leftarrow \cos(z_{\mathrm{sem}}^q,z_{\mathrm{sem}}^r)$
\STATE $\mathcal{M}_{\mathrm{sem},i} \leftarrow 1-\max_j\cos(Z_{\mathrm{sem},i}^q,Z_{\mathrm{sem},j}^r)$ \hfill \textit{// Eq.~\ref{eq:saliency}}
\STATE $\mathbf z_{\mathrm{fused}} \leftarrow \text{CrossAttnFusion}(Z_{\mathrm{cam}}^q,Z_{\mathrm{sem}}^q)$
  \hfill \textit{// Attention, pooling, gating; Q1 only}
\STATE $\mathbf p \leftarrow \text{softmax}(h_{\mathrm{cls}}(z_{\mathrm{fused}}))$; $\hat y\leftarrow\arg\max_k p_k$
  \hfill \textit{// Classification}
\STATE \textit{// OGR Loop:}
\STATE $\text{obs} \leftarrow \text{VLM.Observe}(\xq, \xref)$
\STATE $\text{draft} \leftarrow \text{VLM.Generate}(\text{obs})$
  \hfill \textit{// Initial explanation}
\STATE $\text{output} \leftarrow \text{VLM.Refine}(\text{draft}, \scam, \ssem, \mathcal{M}_{\mathrm{sem}})$
  \hfill \textit{// Score-guided refinement}
\RETURN $\hat{y}$, output, $\mathcal{M}_{\mathrm{sem}}$
\end{algorithmic}
\end{algorithm}


\subsection{VLM Evaluation Setup Details}
\label{app:vlm_eval_details}

This section describes the VLM evaluation metrics used in Section~\ref{sec:vlm_eval}. The VLM output examples and formats can refer to Appendix~\ref{app:qual_extended}.

\subsubsection{Change Type Classification}
We measure three-class accuracy over tampered, retouched, and augmented. Ground-truth addition/removal/replacement/multiple labels map to tampered, retouching to retouched, and augmentation to augmented.

\subsubsection{Object Recall}

Object recall measures how completely the VLM identifies changed objects described in the ground truth. We evaluate this metric on tampered pairs only. Specifically, our computation follows the following steps:

\begin{enumerate}[leftmargin=1.5em,itemsep=1pt]
    \item Extract GT object nouns from the annotation by tokenizing, stemming (plurals, -ing forms), and filtering stopwords, adjectives, colors, and positional words.
    \item Concatenate all predicted \texttt{object\_changes[].description} strings.
    \item Match GT nouns to predictions using substring matching, stemming, and a synonym dictionary (e.g., man/woman $\rightarrow$ person, sofa $\rightarrow$ couch).
    \item Compute: $\text{ObjRecall} = {\text{(\# GT nouns found)}}/{\text{(\# total GT nouns})}$.
\end{enumerate}
We report the averaged Recall across all tampered pairs.

\subsubsection{Object Hit}

It is a stricter object detection metric requiring both \textit{sufficient noun overlap} and \textit{action type agreement}. Evaluated on tampered pairs only.

\noindent\textbf{Requirements for a hit:}
\begin{enumerate}[leftmargin=1.5em,itemsep=1pt]
    \item \textit{Noun threshold}: At least $\min(2, |\text{GT nouns}|)$ GT nouns must appear in predictions.
    \item \textit{Action match}: The predicted action type must be compatible with the GT action type.
\end{enumerate}

\noindent\textbf{Action compatibility rules:}
\begin{itemize}[leftmargin=1.5em,itemsep=1pt]
    \item Same action (addition/removal/replacement) $\rightarrow$ match.
    \item Replacement is compatible with both addition and removal.
    \item If either action is undetectable $\rightarrow$ match (no penalty).
    \item Addition vs.\ removal (without replacement) $\rightarrow$ no match.
\end{itemize}

\subsubsection{Localization Proxy}

Localization is evaluated against proxy GT boxes parsed from free-text annotations on tampered pairs with parseable locations. These coarse spatial targets are separate from mask-derived boxes and pixel-level masks.

\noindent\textbf{Location parsing.} Free-text location descriptions (e.g., ``bottom-right'', ``left side'') are mapped to approximate bboxes on a 0-1000 coordinate system:
\begin{itemize}[leftmargin=1.5em,itemsep=1pt]
    \item Horizontal: \texttt{left} $\rightarrow$ [0, 500], \texttt{right} $\rightarrow$ [500, 1000], \texttt{center} $\rightarrow$ [250, 750].
    \item Vertical: \texttt{top/upper} $\rightarrow$ [0, 500], \texttt{bottom/lower/foreground} $\rightarrow$ [500, 1000].
    \item Corners are tightened to one-third regions (e.g., ``top-left corner'' $\rightarrow$ [0, 0, 333, 333]).
\end{itemize}
Object-relative descriptions (e.g., ``on the dog's face'') and ``entire image'' are skipped.

\subsection{Policy Review under a Mock Policy}
\label{app:policy}

\paragraph{Task and evaluation unit.}
Policy review converts structured forensic evidence into an accept/reject decision under a predefined mock content policy. The evaluation unit is a query--reference image pair, not an individual object or bounding box. We evaluate 400 pairs with human-annotated policy verdicts: 365 accept and 35 reject. These verdicts are distinct from the three modification-type labels: an edit can be correctly identified as tampering yet remain acceptable under the mock policy if it does not violate a protected-content rule. The checker operates on the VLM's structured description of the change, rather than directly assigning a policy verdict from the channel scores.

\paragraph{Mock policy.}
The policy specifies two rejection criteria and permits other edits when neither criterion is triggered:
\begin{itemize}[leftmargin=1.5em, itemsep=2pt, topsep=4pt]
    \item \textbf{Biometric modification (reject):} Alterations to physical identity markers, including skin tone, eye color, hair texture, facial features, or facial geometry.
    \item \textbf{Sensitive entity modification (reject):} Addition, removal, or replacement of people or sensitive entities, including weapons, illegal substances, and hate symbols.
    \item \textbf{Permitted edits (accept):} Aesthetic adjustments (such as retouching, compression, and camera-parameter restyling), changes to animals or ordinary inanimate objects, and background modifications, provided they do not satisfy either rejection criterion.
\end{itemize}
These categories define a task-specific mock policy, rather than a universal definition of acceptable photographic editing.

\paragraph{Baseline and OGR inputs to the checker.}
For each VLM, the baseline supplies the structured output from its image-pair assessment. The +OGR condition supplies the assessment after the Observe--Generate--Refine procedure. The policy checker uses the same rule set for both conditions. It reads the \texttt{identity\_biometric} field and a list of \texttt{object\_changes[]} entries. Each object entry contains a \texttt{change\_type} and a textual \texttt{description}. For +OGR, the checker uses the revised object-change list in \texttt{revised\_semantic\_alignment} when present; otherwise it retains the original list. The biometric field is unchanged between passes. Thus, differences between the two conditions arise through the evidence supplied to the checker, particularly revisions to the described object changes, rather than a different policy.

\paragraph{Rule application and verdict aggregation.}
The checker applies regular-expression matching in two branches:
\begin{enumerate}[leftmargin=1.5em,itemsep=2pt]
    \item \textbf{Biometric branch.} Search \texttt{identity\_biometric} for change phrases such as ``facial geometry changed'', ``identity swap'', and ``face morph''. Safe phrases such as ``no identity manipulation detected'' and ``facial geometry is consistent'' override the biometric trigger. These are examples of the documented patterns, not an exhaustive pattern list.
    \item \textbf{Sensitive-entity branch.} Examine object entries whose \texttt{change\_type} is addition, removal, or replacement. Match their descriptions against a person lexicon (29 terms, including person, man, woman, child, and pedestrian) and a sensitive-entity lexicon (14 terms, including weapon, gun, knife, drug, and hate symbol). A matching eligible entry triggers rejection. Mere mention of an entity in an entry with an ineligible change type does not trigger this branch.
    \item \textbf{Pair-level verdict.} Reject the pair if either branch triggers; otherwise accept it. Multiple flagged objects still produce one pair-level rejection. A safe phrase in the biometric branch does not cancel a sensitive-entity violation.
\end{enumerate}
Writing $b_i$ for a biometric trigger after safe-phrase handling and $s_{ij}$ for an eligible sensitive-entity trigger in object entry $j$, the decision is
\[
\widehat{y}_i = \mathbb{1}\!\left[b_i \lor \bigvee_j s_{ij}\right],
\qquad 1=\text{reject},\quad 0=\text{accept}.
\]
For illustration, an entry reporting the removal of a person triggers the sensitive-entity branch, whereas a background color adjustment without a protected-content change is accepted. A reported face morph triggers the biometric branch even if no object is added or removed. These examples illustrate the decision rules and are not additional experimental observations.

\paragraph{Scoring against human verdicts.}
Each predicted verdict is compared with the human accept/reject label, with reject as the positive class. A true positive is a correctly rejected pair; a false positive is an acceptable pair rejected by the checker; a false negative is a policy-violating pair accepted by the checker; and a true negative is a correctly accepted pair. The pair-level metrics are
\[
\begin{aligned}
\mathrm{Accuracy} &= \frac{TP+TN}{TP+TN+FP+FN}, &
\mathrm{Precision} &= \frac{TP}{TP+FP},\\
\mathrm{Recall} &= \frac{TP}{TP+FN}, &
\mathrm{FPR} &= \frac{FP}{FP+TN}.
\end{aligned}
\]
Table~\ref{tab:ogr-main-results} reports accuracy, precision, and recall, multiplied by 100. Precision measures the reliability of a rejection, while recall measures coverage of policy-violating pairs. This evaluation assesses the combined VLM-description and rule-checker pipeline: an incorrect final verdict may arise from a missing or hallucinated change description, or from the keyword rules themselves.

\paragraph{Results and scope.}
Policy results and their interpretation appear in Table~\ref{tab:ogr-main-results} and Sec.~\ref{sec:vlm_eval}. The policy columns retain the reported point estimates from the policy evaluation, whereas the forensic-evidence and revision-frequency columns use the three-run aggregates also reported in Appendix~\ref{app:ogr_ablation}. Run-level uncertainty estimates are not available for the policy results. Missed or hallucinated entities in the VLM output and limitations of the rule-based checker can both affect the final verdict.

\section{Additional Experiments}
\label{app:exp}

\subsection{Within-Source Classification Diagnostics}
\label{app:within_source}
Table~\ref{tab:within_source} reports two comparisons restricted to a named source collection. This restriction removes variation in the collection name within each comparison, but does not necessarily match underlying original-photo sources, compression, resolution, or generator signatures across classes. The SID-Set AUC indicates ranking discrimination within that collection. The Defactify accuracy additionally depends on class proportions and the decision rule. These metrics therefore cannot be averaged or interpreted as proof that source confounding is absent. Neither comparison includes the GP retouching class.

\begin{table}[htbp]
\centering
\caption{\textbf{Within-source diagnostics.} Values are retained in the metric originally reported for each comparison. These are not matched-original or independent-generator evaluations.}
\label{tab:within_source}
\small
\begin{tabular}{llll}
\toprule
Collection & Comparison & Metric & Result \\
\midrule
SID-Set & Camera vs. tampered & AUC & 95.5\% \\
Defactify & Camera vs. fully synthetic & Accuracy & 90.4\% \\
\bottomrule
\end{tabular}
\end{table}

\subsection{Cross-Pipeline Retouching Diagnostic}
\label{app:rebuttal_transfer}

The OpenCV data are used only for evaluation and are mapped exclusively to the GP evaluation subset; they are not used in either training stage. Table~\ref{tab:retouch_transfer} reports the predicted-class distributions for GP effects and corresponding effects recreated using classical OpenCV operations on the associated evaluation source images. This establishes the intended evaluation split; auditing underlying parent identities across other source collections is a separate check. The OpenCV outputs are assigned the retouched label in 81.2-92.7\% of cases. This is evidence of sensitivity to related visual processing beyond the GP implementation; it is not a test of attribution to GenAI, because the OpenCV operations are non-neural. It also does not replace evaluation on an entirely unseen GenAI retouching tool. Under the GenAI-specific retouching definition, a conventional operation alone does not establish membership in the GenAI-retouched class, regardless of its visual strength. The predictions therefore expose a potential provenance ambiguity. Operation severity and the policy for conventional edits must be specified before this diagnostic can be scored as a classification task.

\begin{table}[htbp]
\centering
\caption{\textbf{Cross-pipeline diagnostic.} Row-normalized predicted-class percentages. ``Retouched'' is a model prediction, not proof of GenAI processing or an established correctness label for OpenCV operations.}
\label{tab:retouch_transfer}
\small
\begin{tabular}{llrrrr}
\toprule
Pipeline & Effect & Camera & Retouched & Tampered & Synthetic \\
\midrule
GP & Color temperature & 1.0 & 97.0 & 2.0 & 0.0 \\
GP & Bokeh & 1.5 & 94.5 & 4.0 & 0.0 \\
GP & Focal length & 5.5 & 84.5 & 10.0 & 0.0 \\
GP & Shutter speed & 1.0 & 97.0 & 1.5 & 0.5 \\
\midrule
OpenCV & Color temperature & 3.2 & 90.6 & 5.4 & 0.8 \\
OpenCV & Bokeh & 2.7 & 81.2 & 15.3 & 0.8 \\
OpenCV & Focal length & 4.4 & 90.0 & 5.4 & 0.2 \\
OpenCV & Shutter speed & 0.3 & 92.7 & 6.7 & 0.3 \\
\bottomrule
\end{tabular}
\end{table}

\subsection{Mild Benign-Processing Diagnostic}
\label{app:benign_diagnostic}

Table~\ref{tab:benign_diagnostic} reports a limited diagnostic under mild settings. Camera recall decreases by at most 4.0 points relative to the unprocessed images. However, the unprocessed camera recall is only 62.5\%, corresponding to a 37.5\% non-camera prediction rate before any transformation; denoising raises that rate to 41.5\%. Small relative changes therefore do not establish a low false-positive rate. This diagnostic does not cover operation severity, real smartphone capture pipelines, or compound processing.

\begin{table}[htbp]
\centering
\caption{\textbf{Mild benign-processing diagnostic.} Predicted-class percentages on camera-derived images. All rows should retain the camera label under this diagnostic's policy.}
\label{tab:benign_diagnostic}
\small
\begin{tabular}{lrrrr}
\toprule
Transformation & Camera & Tampered & Synthetic & Retouched \\
\midrule
No processing & 62.5 & 26.5 & 9.5 & 1.5 \\
Resize & 60.5 & 29.0 & 9.0 & 1.5 \\
Denoise & 58.5 & 29.5 & 10.0 & 2.0 \\
HDR tonemap & 59.5 & 27.5 & 11.0 & 2.0 \\
Social recompress & 60.5 & 27.0 & 11.0 & 1.5 \\
Computational photo & 62.0 & 26.0 & 10.5 & 1.5 \\
\bottomrule
\end{tabular}
\end{table}

\subsection{Per-Generator Classification Results}
\label{app:binary_detail}

Table~\ref{tab:baselines_two_datasets_auc} provides a per-generator breakdown of camera-vs-edited AUC across the Defactify (in-distribution) and Flickr8K$^*$ (held-out) datasets, complementing the dataset-level summary reported elsewhere. \method{} achieves the best AUC on every Defactify generator (.892-.923), showing strong within-source performance across the listed diffusion families (SD2.1/SDXL/SD3), DALL-E~3, and Midjourney rather than overfitting to any one generator. On the held-out Flickr8K$^*$ set, FatFormer leads on all five generators while \method{} remains competitive (.772-.929), and the two methods are essentially tied on the ten-column average (\method{}: .893; FatFormer: .894).

\begin{table}[h]
\centering
\caption{\textbf{Camera authenticity across generators (AUC).} Defactify and Flickr8K*, each with five generators. This breakdown covers eight baselines; SAFE and GAPL are included only in the main four-class comparison. \textit{Flickr8K*} is held-out. Best in \textbf{bold}, second \underline{underlined}.}
\label{tab:baselines_two_datasets_auc}
\setlength{\tabcolsep}{4.2pt}
\renewcommand{\arraystretch}{1.05}
\resizebox{\linewidth}{!}{%
\begin{tabular}{lccccccccccc}
\toprule
\textbf{Method / AUC} &
\multicolumn{5}{c}{\textbf{Defactify Images}} &
\multicolumn{5}{c}{\textbf{Flickr8K*}} &
\textbf{Avg.} \\
\cmidrule(lr){2-6}\cmidrule(lr){7-11}
& \textbf{SD2.1} & \textbf{SDXL} & \textbf{SD3} & \textbf{DALL-E 3} & \textbf{Midjourney 6}
& \textbf{SDAbso.} & \textbf{SDEpic.} & \textbf{SDv1.5} & \textbf{SDv6.0} & \textbf{SDXL} & \\
\midrule
TruFor~\citep{guillaro2023trufor} & .455 & .407 & .382 & .452 & .516 & .656 & .668 & .660 & .550 & .689 & .544 \\
D3~\citep{yang2025d} & .613 & .715 & .618 & .778 & .762 & .723 & .693 & .668 & .643 & .635 & .685 \\
CNNSpot~\citep{wang2020cnn} & .561 & .547 & .542 & .606 & .560 & .625 & .574 & .598 & .550 & .533 & .570 \\
AIDE~\citep{yan2024sanity} & .626 & .701 & .633 & .567 & \underline{.862} & .929 & .905 & .895 & .835 & .796 & .775 \\
VIBAIGCD~\citep{zhang2025towards} & .757 & .808 & .740 & .844 & .820 & \underline{.932} & \underline{.924} & \underline{.922} & \underline{.903} & .698 & .835 \\
FreqNet~\citep{tan2024frequency} & .541 & .584 & .522 & .544 & .579 & .851 & .868 & .837 & .843 & \underline{.823} & .699 \\
FatFormer~\citep{liu2024forgery} & \underline{.836} & \underline{.860} & \underline{.812} & \underline{.895} & .839 & \textbf{.967} & \textbf{.962} & \textbf{.961} & \textbf{.958} & \textbf{.850} & \textbf{.894} \\
UnivFD~\citep{ojha2023towards} & .751 & .781 & .752 & .852 & .805 & .876 & .861 & .852 & .845 & .676 & .805 \\
\midrule
\textbf{\method{}}
  & \textbf{.895} & \textbf{.899} & \textbf{.892} & \textbf{.923} & \textbf{.911}
  & .929 & .905 & .910 & .893 & .772 & \underline{.893} \\
\bottomrule
\end{tabular}%
}
\end{table}

\subsection{Matched-source Four-class Diagnostic}
\label{app:matched_source_diagnostic}

We additionally test whether four-way predictions remain reliable when all four labels are derived from the same source scene. We select 100 original-retouched pairs from the existing Generative Photography test-source pool, with 25 pairs per camera setting (bokeh, color temperature, focal length, and shutter speed). The camera and retouched images are the existing paired files. For each camera original, Big-LaMa removes a masked object to form a tampered candidate, with pixels outside the mask copied from the original. Separately, FLUX.1-dev generates a fully synthetic image from the original's caption alone, without image conditioning. This produces 100 source groups and 400 images (100 per class).

We evaluate frozen Stage A and Stage B checkpoints without retraining or tuning on these images. Camera and SigLIP2 patch-token embeddings are extracted once and supplied to the full model and its camera-only and semantic-only ablations. Each model receives one image at a time, without access to the group identity. Table~\ref{tab:matched_source_diagnostic} reports four-way argmax accuracy, macro-F1, macro one-vs-rest AUC, and class recalls. The full model attains 46.25\% accuracy, compared with 44.50\% for semantic-only. Its 1.75-point accuracy difference has a paired 95\% bootstrap interval of $[-0.75,4.50]$ points when resampling the 100 original-image groups, and therefore does not establish a reliable gain over the semantic channel alone. In particular, the full model identifies only 4\% of camera images and 8\% of tampered candidates.

\begin{table}[htbp]
\centering
\caption{\textbf{Matched-source four-class diagnostic.} All values are percentages.}
\label{tab:matched_source_diagnostic}
\small
\begin{tabular}{@{}lrrr@{}}
\toprule
\textbf{Frozen model} & \textbf{Acc.} & \textbf{Macro-F1} & \textbf{Macro-AUC} \\
\midrule
Full \method{} & 46.25 & 39.14 & 71.99 \\
Camera only & 28.75 & 23.88 & 64.33 \\
Semantic only & 44.50 & 36.51 & 66.42 \\
\bottomrule
\end{tabular}
\end{table}

The full model's recalls are 4\%, 99\%, 8\%, and 74\% for camera, retouched, tampered, and fully synthetic images, respectively; the corresponding per-class one-vs-rest AUCs are 0.450, 0.848, 0.582, and 0.999. These ranking scores should not obscure the poor camera and tampered argmax recalls. The cohort reuses the paper's existing test-source pool, rather than providing a new independent model-selection holdout. Exact source-path and pixel-change checks passed, but visual review of all Big-LaMa removals and scene consistency is incomplete; some spot-checked removals are small or visibly blurred. We therefore treat this as a failure-oriented diagnostic, not evidence that matched-source confounding has been resolved. The group manifest, model-specific generation records, predictions, and inference logs are retained with the dataset for auditing.

\subsection{Validation of VLM Evaluation with Annotated Bounding Box}
\label{app:proxy_gt}

In Sec.~\ref{sec:experiments} of the main paper, the VLM evidence evaluation uses
location text as a proxy ground truth for localization quality (IoU) as not all the datasets have bounding-box annotation.
To validate this proxy, we manually annotate bounding boxes for a  subset of tampered image pairs from SAGI-D (those for which the VLM actually produces a bounding box), using the ground-truth masks provided by the dataset. We then compare the VLM-generated bounding boxes against both the human-annotated boxes and the saliency-derived proxy boxes.

\begin{table}[h]
\centering
\caption{\textbf{Proxy GT validation.} Agreement between VLM-derived bounding boxes and human-annotated bounding boxes on 50 tampered pairs.}
\label{tab:proxy_validation}
\begin{tabular}{@{}lcccc@{}}
\toprule
& \multicolumn{2}{c}{Baseline} & \multicolumn{2}{c}{Ours} \\
\cmidrule(lr){2-3} \cmidrule(lr){4-5}
Metric & Mean & Std & Mean & Std \\
\midrule
IoU        & .500 & .241 & .597 & .228 \\
Precision  & .722 & .244 & .823 & .203 \\
Recall     & .638 & .297 & .690 & .254 \\
\bottomrule
\end{tabular}
\end{table}

Table~\ref{tab:proxy_validation} reports higher agreement with mask-derived human boxes for OGR on the selected 50-pair subset (IoU .500 to .597). This supports localization on that subset, but does not directly validate the text-derived proxy: doing so requires comparing proxy boxes themselves with mask-derived boxes. Selecting pairs on which a VLM outputs a box can also exclude localization failures. The subset selection and missing-box handling therefore limit the conclusion.

\subsection{Design-Space Comparison with VLM-Based Forensic Methods}
\label{app:vlm_baselines}

Table~\ref{tab:vlm_designspace} provides a systematic design-space comparison with VLM-based forensic methods, clarifying fundamental
architectural and formulation differences.

\paragraph{Scope of a direct comparison.}
Forensic VLMs differ in reference access, training data, and output format. These differences require explicit reporting; they do not make practical comparison infeasible. Our current experiments measure improvements to the same frozen VLM under pairwise inputs, rather than competitiveness against fine-tuned forensic systems. A complementary comparison could evaluate a released forensic VLM on a shared camera/tampered subset with mask-derived boxes, showing its native single-image results separately from any pair-input adaptation and reporting the reference advantage of OGR.

\begin{table}[h]
\centering
\caption{\textbf{Design-space comparison with VLM-based forensic methods.}
  \method{} is the only method that (i)~uses reference-based verification,
  (ii)~decomposes camera and semantic channels, (iii)~distinguishes
  retouching from tampering, and (iv)~uses the tested VLMs without forensic instruction tuning.}
\label{tab:vlm_designspace}
\footnotesize
\resizebox{\linewidth}{!}{%
\begin{tabular}{@{}lccccc@{}}
\toprule
\textbf{Property} & \textbf{FakeVLM} & \textbf{FakeShield} & \textbf{ForgeryGPT} & \textbf{SIDA} & \textbf{\method{}} \\
\midrule
Input type & Single & Single & Single & Single & \textbf{Pair} \\
VLM training & SFT & SFT & Staged tuning & SFT & \textbf{Frozen} \\
Channel decomp. & \texttimes & \texttimes & \texttimes & \texttimes & \textbf{\checkmark} \\
Retouch vs.\ tamp. & \texttimes & \texttimes & \texttimes & \texttimes & \textbf{\checkmark} \\
Forgery localization & \texttimes & Mask & Mask & Mask & \textbf{Saliency} \\
Output type & Binary+text & Binary+mask & Desc.+mask & Label+mask+desc. & \textbf{Scores+sal.+OGR} \\
VLM portability & Fixed & Fixed & Fixed & Fixed & \textbf{Tested VLMs} \\
\bottomrule
\end{tabular}%
}
\end{table}

\paragraph{Key advantage.} The OGR interface uses frozen VLMs without forensic instruction tuning, while the camera encoder and fusion classifier are trained. The experiments demonstrate improvements across four tested proprietary configurations from the GPT-4o and Gemini~2.5 families. Extension to other backends is plausible but requires evaluation, and the current results do not establish a performance ranking against fine-tuned forensic VLMs.

\subsection{Discussion of Reference-Based Baselines}
\label{app:ref_baselines}

We consider the following reference-based similarity alternatives for change assessment (Sec.~\ref{sec:vlm_eval}):

\begin{itemize}[nosep]
  \item \textbf{C2PA/Content Credentials:} Cryptographic provenance, not learned forensic analysis. Requires adoption at capture time and does not work for legacy content.
  \item \textbf{Perceptual hashing (pHash)~\citep{zauner2010implementation}:} Detects near-duplicates but cannot characterize \textit{what} changed or distinguish retouching from tampering.
  \item \textbf{Image-pair similarity (LPIPS~\citep{zhang2018unreasonable}, DISTS~\citep{ding2020image}):} Measures perceptual distance but provides no forensic decomposition into change type, affected objects, or spatial localization.
\end{itemize}

A quantitative comparison with these reference-based alternatives requires the same evaluation pairs, a specified multi-class decision rule, and thresholds selected on a separate validation split. We leave this matched comparison to further evaluation.

\clearpage
\subsection{OGR Component Ablation}
\label{app:ogr_ablation}

Table~\ref{tab:ogr-ablation} evaluates the contribution of each forensic input over three runs on the 400-pair evaluation set. Each ablation retains OGR while removing the semantic score, camera score, or saliency heatmap. The baseline and full-OGR means and standard deviations are the same as those in the forensic-evidence and revision-frequency columns of Table~\ref{tab:ogr-main-results}. The ablation rows remove one input at a time under the same reporting protocol. Policy-review metrics are not included in this ablation.

\begin{table}[htbp]
\centering\small
\newcommand{\ogrstat}[2]{\ensuremath{#1}\,{\scriptsize\color{black!65}\ensuremath{\pm#2}}}
\setlength{\tabcolsep}{4pt}
\renewcommand{\arraystretch}{1.10}
\caption{\textbf{VLM evidence generation and component ablation.} All values are percentages; smaller gray values denote $\pm$standard deviation over three runs. Shaded rows: full OGR; indented rows: removal of one OGR input. Bold marks the highest mean per model (including ties). $\uparrow$: higher is better. IoU is a localization proxy; revision frequency (Rev) has no preferred direction.}
\label{tab:ogr-ablation}
\begin{tabularx}{\linewidth}{l*{5}{>{\centering\arraybackslash}X}}
\toprule
\textbf{Configuration} & \textbf{ChgType} $\uparrow$ & \textbf{Recall} $\uparrow$ & \textbf{Hit} $\uparrow$ & \textbf{IoU} $\uparrow$ & \textbf{Rev (\%)} \\
\midrule
\multicolumn{6}{l}{\textbf{GPT-4o-mini}} \\
Baseline & \ogrstat{66.1}{0.5} & \ogrstat{21.3}{0.9} & \ogrstat{32.5}{1.1} & \ogrstat{25.6}{0.6} & -- \\
\rowcolor{black!5}
\textbf{Ours (+OGR)} & \ogrstat{72.1}{0.5} & \ogrstat{\mathbf{29.3}}{0.6} & \ogrstat{\mathbf{39.7}}{1.6} & \ogrstat{\mathbf{36.1}}{1.5} & \ogrstat{76.4}{0.3} \\
\hspace{0.7em}w/o semantic & \ogrstat{67.2}{1.6} & \ogrstat{27.7}{0.5} & \ogrstat{38.2}{1.1} & \ogrstat{32.8}{0.0} & \ogrstat{71.3}{2.8} \\
\hspace{0.7em}w/o camera & \ogrstat{69.8}{0.6} & \ogrstat{28.4}{0.6} & \ogrstat{38.0}{1.6} & \ogrstat{35.4}{0.3} & \ogrstat{71.5}{0.0} \\
\hspace{0.7em}w/o heatmap & \ogrstat{\mathbf{72.7}}{0.6} & \ogrstat{25.5}{0.6} & \ogrstat{35.5}{0.8} & \ogrstat{34.3}{1.2} & \ogrstat{69.2}{0.3} \\
\midrule
\multicolumn{6}{l}{\textbf{GPT-4o}} \\
Baseline & \ogrstat{84.4}{0.3} & \ogrstat{35.6}{0.7} & \ogrstat{59.4}{1.2} & \ogrstat{60.9}{3.7} & -- \\
\rowcolor{black!5}
\textbf{Ours (+OGR)} & \ogrstat{\mathbf{89.8}}{0.2} & \ogrstat{36.4}{0.2} & \ogrstat{60.3}{0.3} & \ogrstat{61.3}{0.3} & \ogrstat{15.2}{0.8} \\
\hspace{0.7em}w/o semantic & \ogrstat{85.7}{0.1} & \ogrstat{\mathbf{36.5}}{0.1} & \ogrstat{\mathbf{60.6}}{0.5} & \ogrstat{\mathbf{61.5}}{0.3} & \ogrstat{11.7}{0.3} \\
\hspace{0.7em}w/o camera & \ogrstat{87.0}{0.7} & \ogrstat{36.0}{0.1} & \ogrstat{60.1}{0.0} & \ogrstat{61.1}{0.0} & \ogrstat{5.1}{0.9} \\
\hspace{0.7em}w/o heatmap & \ogrstat{87.4}{0.1} & \ogrstat{36.0}{0.1} & \ogrstat{60.3}{0.3} & \ogrstat{61.1}{0.0} & \ogrstat{22.8}{0.4} \\
\midrule
\multicolumn{6}{l}{\textbf{Gemini 2.5 Flash}} \\
Baseline & \ogrstat{71.0}{1.0} & \ogrstat{41.8}{2.1} & \ogrstat{64.0}{2.3} & \ogrstat{57.4}{1.4} & -- \\
\rowcolor{black!5}
\textbf{Ours (+OGR)} & \ogrstat{\mathbf{78.7}}{2.6} & \ogrstat{\mathbf{46.8}}{0.5} & \ogrstat{\mathbf{72.4}}{0.3} & \ogrstat{\mathbf{70.0}}{1.7} & \ogrstat{36.8}{1.1} \\
\hspace{0.7em}w/o semantic & \ogrstat{77.6}{0.8} & \ogrstat{46.4}{0.3} & \ogrstat{72.1}{0.3} & \ogrstat{69.3}{1.5} & \ogrstat{39.2}{2.0} \\
\hspace{0.7em}w/o camera & \ogrstat{74.4}{0.8} & \ogrstat{45.0}{0.7} & \ogrstat{70.6}{0.3} & \ogrstat{64.1}{0.8} & \ogrstat{13.6}{0.6} \\
\hspace{0.7em}w/o heatmap & \ogrstat{75.2}{1.0} & \ogrstat{44.9}{0.2} & \ogrstat{70.5}{0.3} & \ogrstat{62.0}{0.3} & \ogrstat{29.2}{1.1} \\
\midrule
\multicolumn{6}{l}{\textbf{Gemini 2.5 Pro}} \\
Baseline & \ogrstat{73.1}{0.8} & \ogrstat{54.0}{0.2} & \ogrstat{77.6}{1.2} & \ogrstat{73.5}{0.8} & -- \\
\rowcolor{black!5}
\textbf{Ours (+OGR)} & \ogrstat{\mathbf{77.8}}{0.6} & \ogrstat{55.3}{0.3} & \ogrstat{\mathbf{79.0}}{0.3} & \ogrstat{\mathbf{76.1}}{0.0} & \ogrstat{16.8}{1.7} \\
\hspace{0.7em}w/o semantic & \ogrstat{77.4}{0.8} & \ogrstat{55.3}{0.5} & \ogrstat{78.3}{1.0} & \ogrstat{75.4}{0.8} & \ogrstat{20.5}{0.5} \\
\hspace{0.7em}w/o camera & \ogrstat{76.2}{0.2} & \ogrstat{\mathbf{55.5}}{0.4} & \ogrstat{78.1}{0.3} & \ogrstat{75.2}{0.6} & \ogrstat{10.9}{0.8} \\
\hspace{0.7em}w/o heatmap & \ogrstat{76.2}{0.2} & \ogrstat{54.8}{0.1} & \ogrstat{78.8}{0.5} & \ogrstat{75.0}{0.0} & \ogrstat{12.0}{0.2} \\
\bottomrule
\end{tabularx}
\end{table}

\noindent\textbf{Contributions of the forensic inputs.} Removing either alignment score lowers mean ChgType accuracy for every model. The semantic score has the larger effect for GPT-4o-mini and GPT-4o, whereas Gemini-Flash and -Pro show the opposite ordering. Thus, the relative utility of the two signals depends on the VLM. The heatmap's contribution is most visible in spatial and object-level metrics for Gemini-Flash and GPT-4o-mini. Full OGR is not uniformly best across all ablations, and the differences vary by model and metric.

\clearpage
\raggedbottom
\section{Additional Qualitative Results}
\label{app:qualitative}

\subsection{Extended Qualitative Examples}
\label{app:qual_extended}

\begin{figure}[H]
    \centering
    \includegraphics[width=\linewidth]{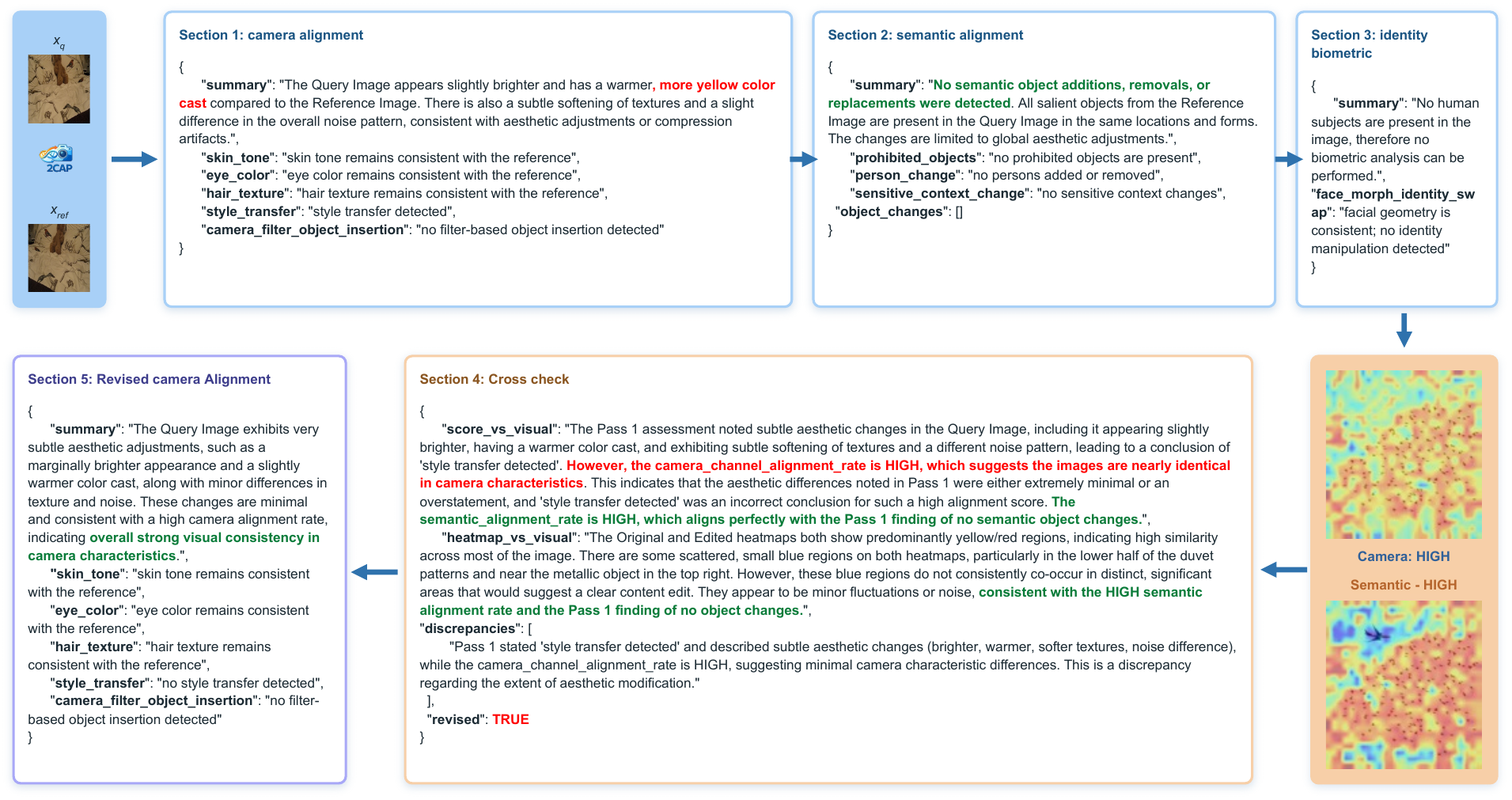}
    \caption{\textbf{OGR example: benign augmentation.} The refinement retracts the initial style-transfer assessment and reports no object changes. The pair illustrates a camera-derived augmentation rather than localized tampering.}
    \label{fig:stage3_example1}
\end{figure}

\begin{figure}[H]
    \centering
    \includegraphics[width=\linewidth]{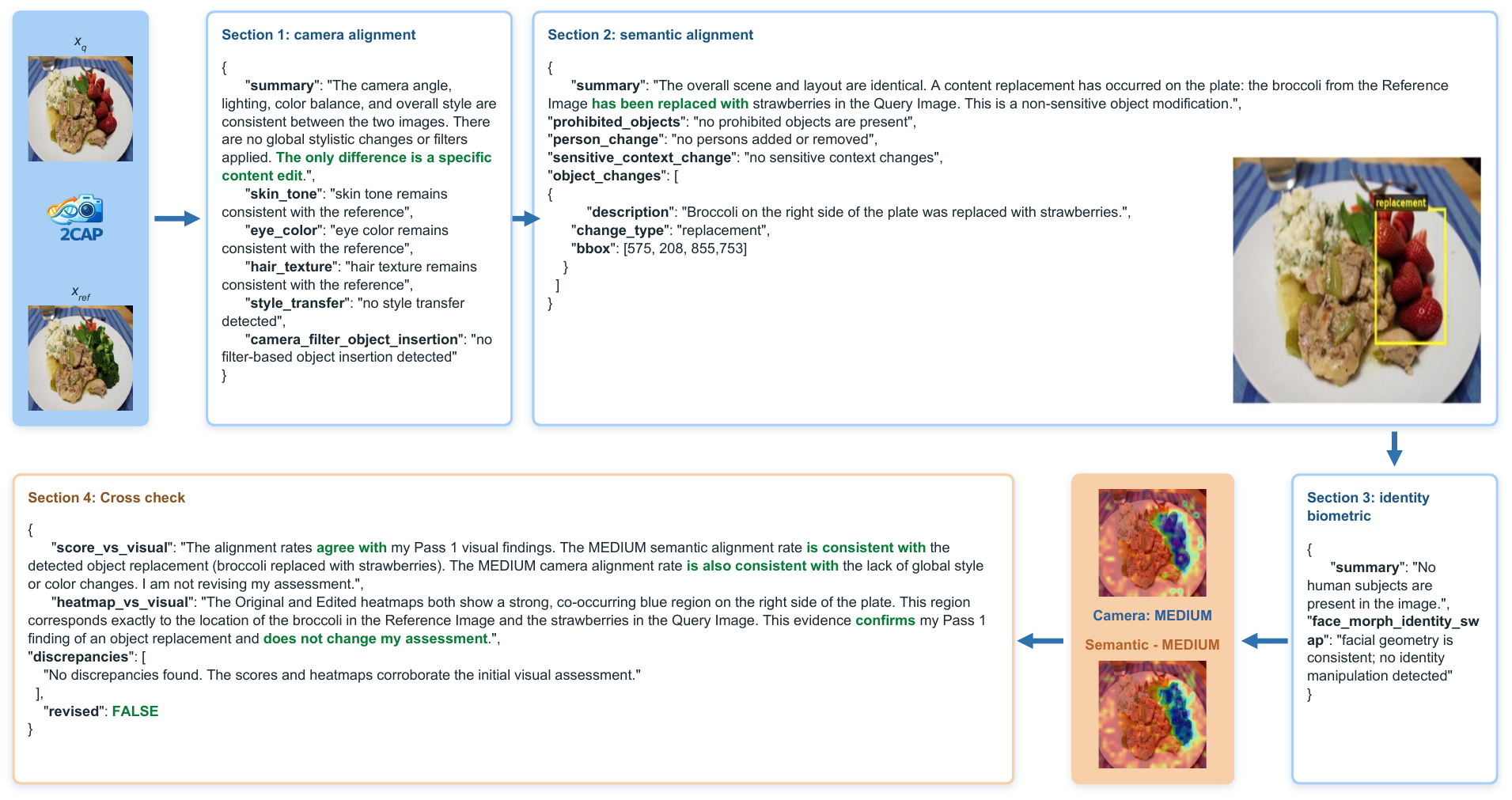}
    \caption{\textbf{OGR example: localized tampering by replacement.} Broccoli on the right side of the plate is replaced with strawberries. The object-change description identifies a replacement rather than a global retouch.}
    \label{fig:stage3_example2}
\end{figure}

\begin{figure}[H]
    \centering
    \includegraphics[width=\linewidth]{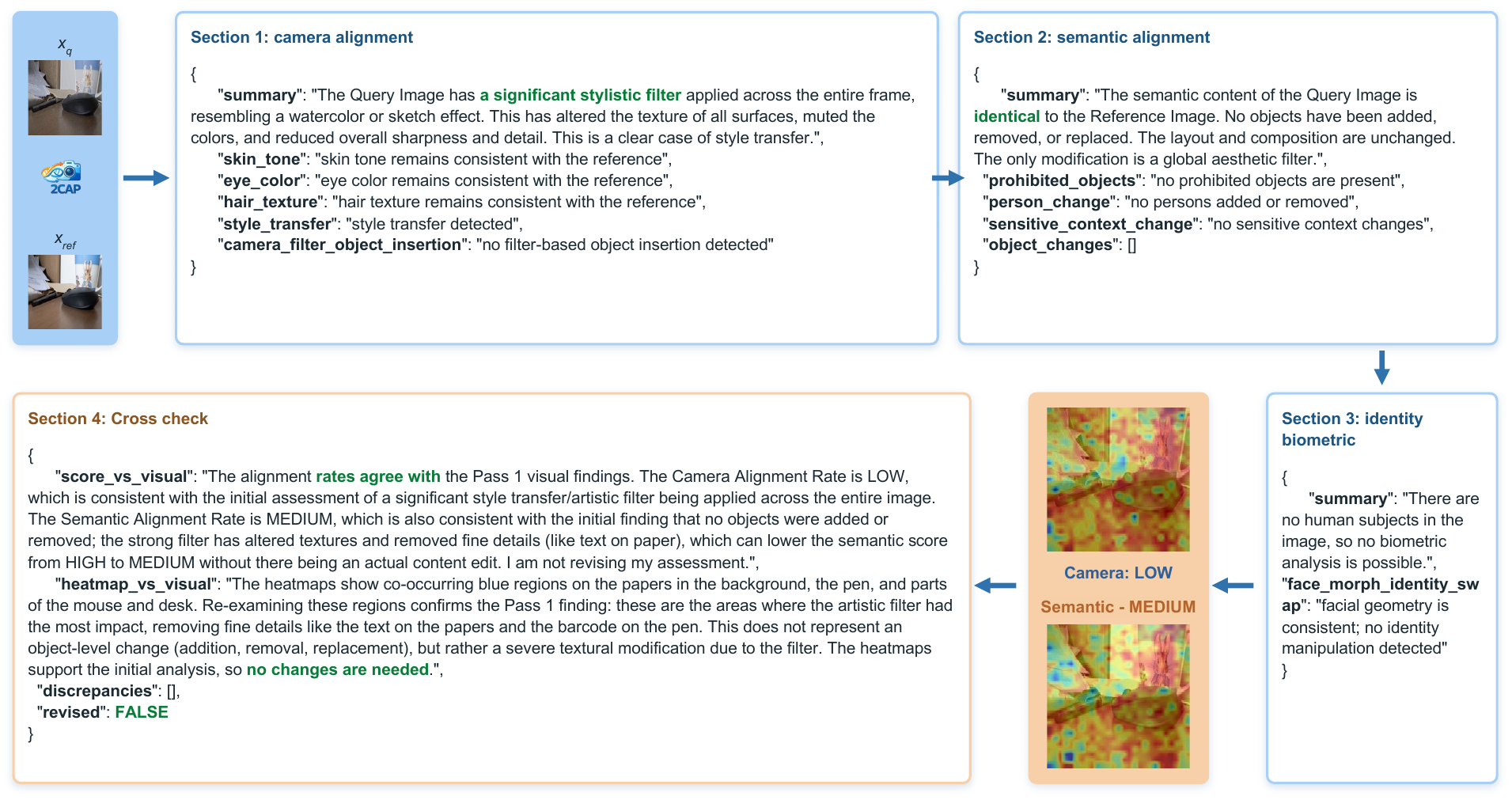}
    \caption{\textbf{OGR example: retouching by style transfer.} A global watercolor/sketch-like aesthetic filter changes appearance while the displayed output reports no object changes. This is a paired retouching example, not text-to-image full synthesis.}
    \label{fig:stage3_example3}
\end{figure}

\begin{figure}[H]
    \centering
    \includegraphics[width=\linewidth]{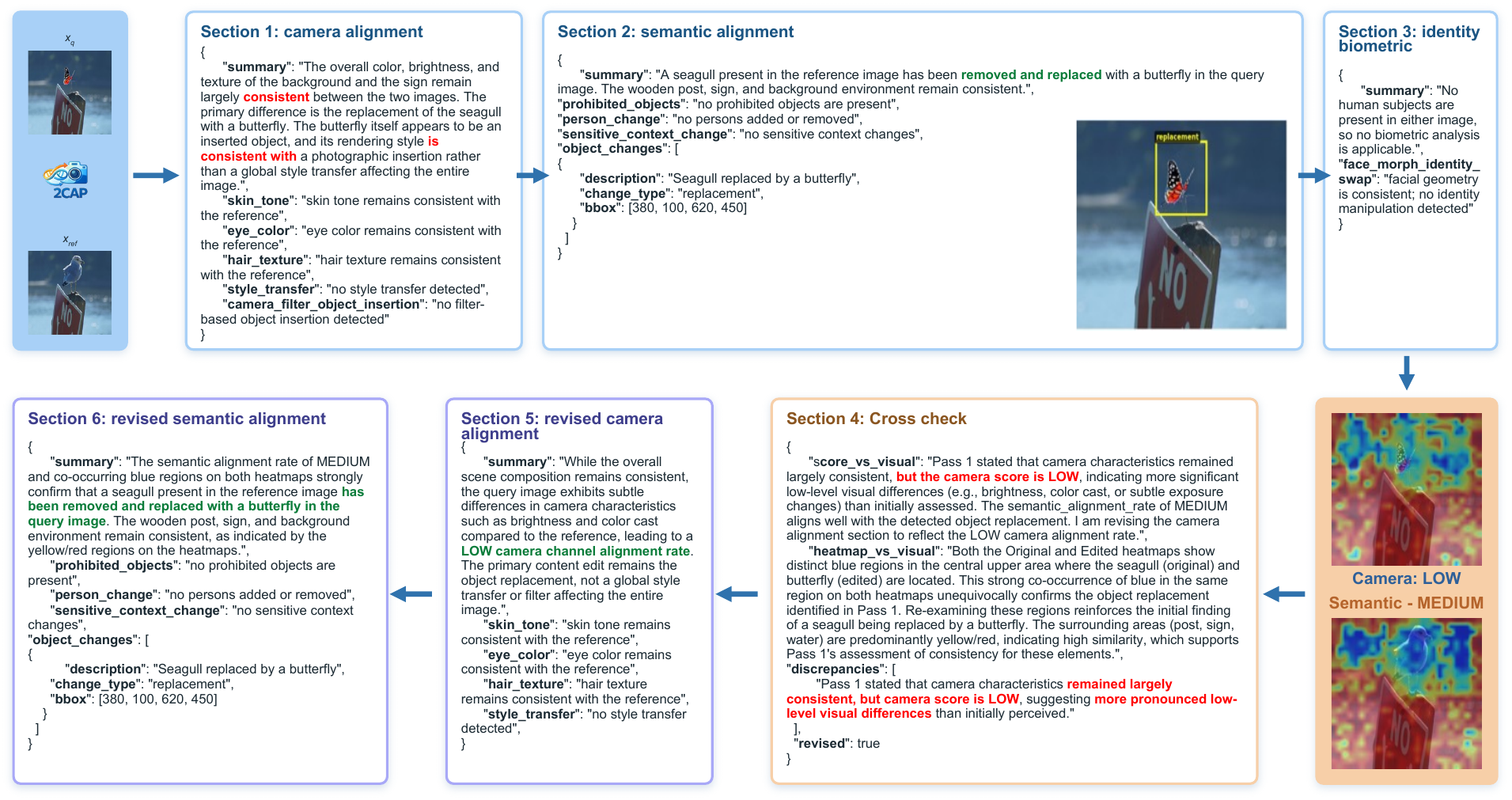}
    \caption{\textbf{OGR example: localized tampering by replacement.} A seagull is replaced with a butterfly. The paired reference makes the changed object and operation explicit.}
    \label{fig:stage3_example4}
\end{figure}

\begin{figure}[H]
    \centering
    \includegraphics[width=\linewidth]{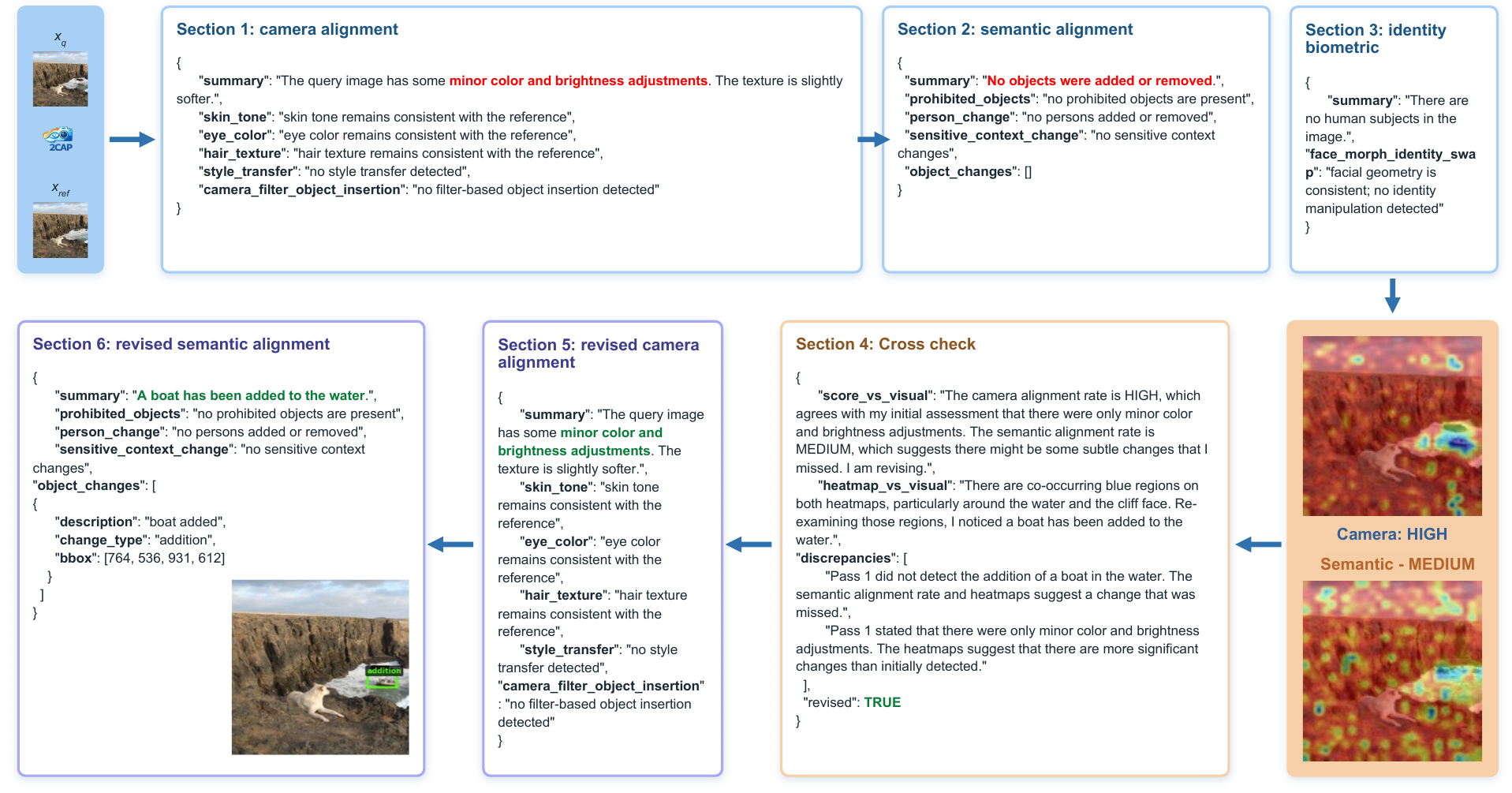}
    \caption{\textbf{OGR example: localized tampering by addition.} A boat is added to the water. The displayed cross-check reports high camera alignment, illustrating that localized tampering can coexist with preserved camera-related cues.}
    \label{fig:stage3_example5}
\end{figure}

Figures~\ref{fig:stage3_example1}-\ref{fig:stage3_example5} present additional qualitative results beyond the main paper, covering benign augmentation, retouching, and localized tampering. Each example shows the query-reference pair, the two-pass OGR pipeline output (Pass~1 visual-only assessment, cross-check, and Pass~2 revision when triggered), and the final forensic conclusion. These examples illustrate the outputs of the Observe--Generate--Refine loop, describing \textit{what} changed, \textit{where}, and \textit{how}. We include both clear-cut cases where the dual-channel signatures are distinctive and borderline cases where scores are ambiguous, illustrating the range of \method{}'s behavior across diverse editing scenarios.

\subsection{Failure Case Analysis}
\label{app:failure}

\begin{figure}[h]
    \centering
    \includegraphics[width=0.9\linewidth]{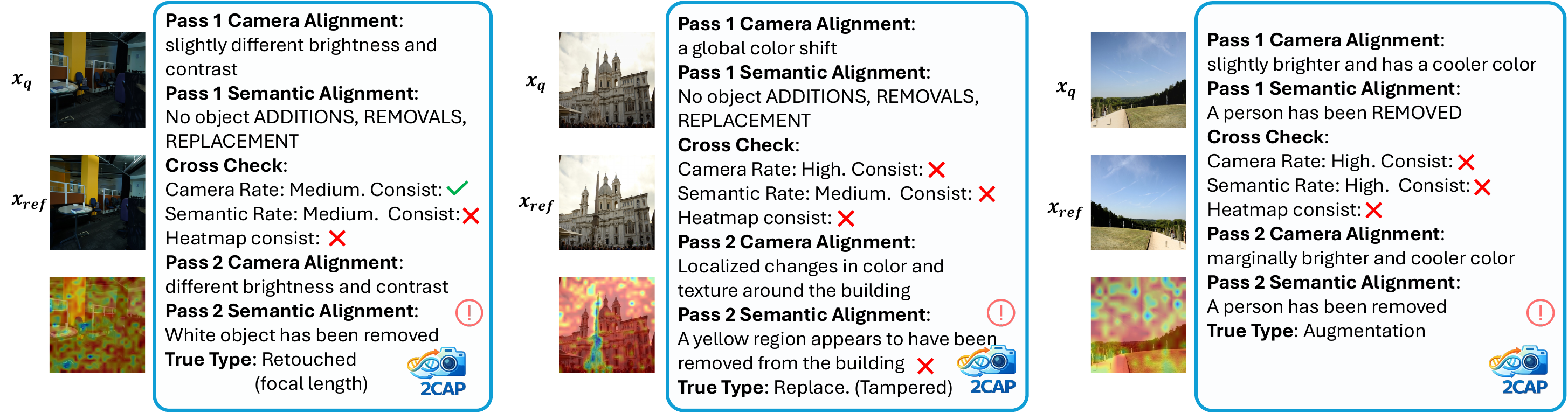}
    \caption{\textbf{Failure cases of \method{}.} Three challenging examples where the two-pass pipeline produces incorrect conclusions. Left: retouching misclassified due to VLM hallucination under ambiguous saliency. Middle: a building-region replacement missed in Pass 1, with Pass 2 incorrectly describing the edit as a removal. Right: augmentation misclassified as tampering due to VLM hallucinating a person removal.}
    \label{fig:fail-case}
\end{figure}

Understanding failure modes is critical for assessing deployment readiness. We present three representative failure cases in Figure~\ref{fig:fail-case}, each selected because it represents a genuinely difficult sample that challenges not only \method{} but also human annotators. These edge cases reveal the boundaries of the current dual-channel framework and inform directions for future improvement.

\paragraph{Case 1: Retouching with hallucinated object change (left).}
The query image is a focal-length retouching of the reference: a global camera-parameter edit that preserves all semantic content. This is an inherently difficult sample because the focal-length change introduces perspective cropping that creates apparent object disappearances at image borders. Pass~1 correctly identifies brightness and contrast differences with no object changes. However, the cross-check flags heatmap inconsistency, and Pass~2 hallucinates that a ``white object has been removed,'' likely triggered by border-region saliency artifacts from the perspective shift. The result is a misclassification from retouched to tampered.

\paragraph{Case 2: Subtle building-region replacement with an imprecise description (middle).}
The query contains a localized replacement affecting a region of the building. Pass~1 describes a global color shift without identifying the localized edit. After the cross-check, Pass~2 detects a change around the building but describes it as a removal rather than a replacement.

\paragraph{Case 3: Augmentation with hallucinated person removal (right).}
The query is an authentic augmentation (camera-parameter restyling) with no semantic changes, yet Pass~1 confidently reports ``a person has been REMOVED.'' The cross-check flags inconsistencies, but Pass~2 reinforces the hallucination rather than correcting it. This failure arises because the VLM's instruction prompt implicitly biases toward finding edits; when saliency maps show diffuse activity (common in augmentations with global brightness/color shifts), the VLM may anchor on spurious visual differences and fabricate a plausible edit narrative.

\paragraph{Discussion and mitigations.}
These failure cases share a common theme: they are \textit{challenging} samples that lie near the decision boundaries between modification categories. Two key observations emerge:

\begin{enumerate}[nosep]
  \item \textbf{VLM hallucination under weak saliency} (Cases 1 and 3): When both channels produce ambiguous scores, saliency maps are diffuse, providing the VLM with insufficient spatial grounding. The +OGR cross-check partially mitigates this--it correctly detects inconsistencies in all three cases--but the current Pass~2 revision cannot fully suppress hallucination when the VLM has high confidence in its fabricated description. One potential mitigation is training a lightweight hallucination detector that cross-references VLM claims against the quantitative channel scores. We leave these extensions to future work.
  \item \textbf{Edit-operation ambiguity} (Case 2): The revised explanation identifies the building region but confuses replacement with removal. A possible mitigation is an explicit comparison of what occupies the region in the reference and query before assigning an operation label; this remains to be evaluated.
\end{enumerate}

\noindent These examples characterize possible failure mechanisms. Aggregate improvements on the 400-pair evaluation do not establish their prevalence in deployment or the frequency of hallucination; that requires explicit failure annotation and measurement.

\clearpage
\flushbottom
\section{Broad Impact and Limitations}
\label{sec:discussion}

\noindent\textbf{Policy-dependent decisions.}
\method{} provides channel similarity scores, localized evidence, and
natural-language explanations to support policy-dependent human review. Different
stakeholders, such as newsroom editors, scientific reviewers, legal
professionals, platform moderators, may apply different thresholds and
rubrics. \method{} is designed to produce structured evidence without assuming a
universal definition of disallowed manipulation; this
policy-agnosticism is a deliberate design choice.

\noindent\textbf{Adversarial robustness.}
Sophisticated adversaries may craft modifications that fool both
channels simultaneously. While we observe robustness to standard perturbations and
common post-processing pipelines, dedicated adversarial attacks that
preserve apparent camera traces while introducing
semantic modifications remain an open challenge for all forensic systems.

\noindent\textbf{Score context influence.}
While the VLM can in principle disagree with the channel scores, we
observe that score context exerts a strong anchoring effect: when
scores are misleading (\textit{e.g.}, a sophisticated edit that
preserves camera traces), the VLM may under-weight its own visual
observations. Presenting scores as context rather than hard labels mitigates but
does not fully eliminate this anchoring bias. Developing score-aware calibration strategies that balance VLM autonomy with quantitative guidance is an important direction.

\noindent\textbf{Semantic encoder scope.}
Our evidence-generation experiments use frozen SigLIP2 for both the semantic score and patch-level saliency map. The component ablations do not measure sensitivity to the choice of semantic encoder, including cases where highly realistic synthetic or edited images confuse the encoder. Comparing multiple frozen semantic encoders and adversarially stress-testing their semantic matches are important next steps.

\noindent\textbf{VLM reliability.}
Generated explanations may occasionally be inaccurate or
hallucinate forensic details not present in the image. The structured evidence trail (dual-channel scores, saliency
maps) provides independent verification beyond the text, and
human-in-the-loop verification remains essential for high-stakes
decisions.


\section{Declaration of LLM Usage}
\label{app:llm_usage}

In addition to incorporating LLMs into the proposed method, where they are part of the pipeline and evaluation, we also used general-purpose LLM-based assistants to help prepare this paper.

\paragraph{Coding assistance.} LLMs were occasionally consulted to help with auxiliary coding tasks such as debugging error messages and improving code readability. All code that produced the experiments, datasets, models, and numerical results reported in this manuscript was subsequently inspected, tested, and validated by the authors. The architecture, training procedure, evaluation protocol, and analysis scripts were designed by the authors.

\paragraph{Language polishing.} LLMs were used as a writing aid to suggest minor revisions for clarity, grammar, and word choice on text that the authors had already drafted. For every LLM-suggested revision that we adopted, the authors carefully read the rephrased text, confirmed that it preserves the intended technical meaning of the original sentence, and edited or rejected the suggestion whenever it introduced any ambiguity, imprecision, or change in meaning. 

\end{document}